\pdfoutput=1
\documentclass[11pt]{article}

\usepackage[final]{acl}

\usepackage{times}
\usepackage{latexsym}
\usepackage{multirow}

\usepackage[T1]{fontenc}
\usepackage[utf8]{inputenc}

\usepackage{microtype}

\usepackage{inconsolata}

\usepackage{graphicx}
\usepackage{subcaption} % for subfigures
\usepackage{xurl}
\usepackage{hyperref}
\usepackage{enumitem}
\usepackage{authblk}

\title{Navigating the Shortcut Maze: A Comprehensive Analysis of Shortcut Learning in Text Classification by Language Models}

\author[1]{Yuqing Zhou}
\author[2]{Ruixiang Tang}
\author[1]{Ziyu Yao}
\author[1]{Ziwei Zhu}
\affil[1]{George Mason University}
\affil[2]{Rutgers University}
\affil[1]{\{yzhou31, ziyuyao, zzhu20\}@gmu.edu}
\affil[2]{ruixiang.tang@rutgers.edu}

\begin{document}
\maketitle
\begin{abstract}
Language models (LMs), despite their advances, often depend on spurious correlations, undermining their accuracy and generalizability. This study addresses the overlooked impact of subtler, more complex shortcuts that compromise model reliability beyond oversimplified shortcuts. We introduce a comprehensive benchmark that categorizes shortcuts into occurrence, style, and concept, aiming to explore the nuanced ways in which these shortcuts influence the performance of LMs. Through extensive experiments across traditional LMs, large language models, and state-of-the-art robust models, our research systematically investigates models' resilience and susceptibilities to sophisticated shortcuts. Our benchmark and code can be found at:
\url{https://github.com/yuqing-zhou/shortcut-learning-in-text-classification}.
\end{abstract}

\section{Introduction}

\begin{figure}[t!]
    \centering
    \includegraphics[width=0.5\textwidth]{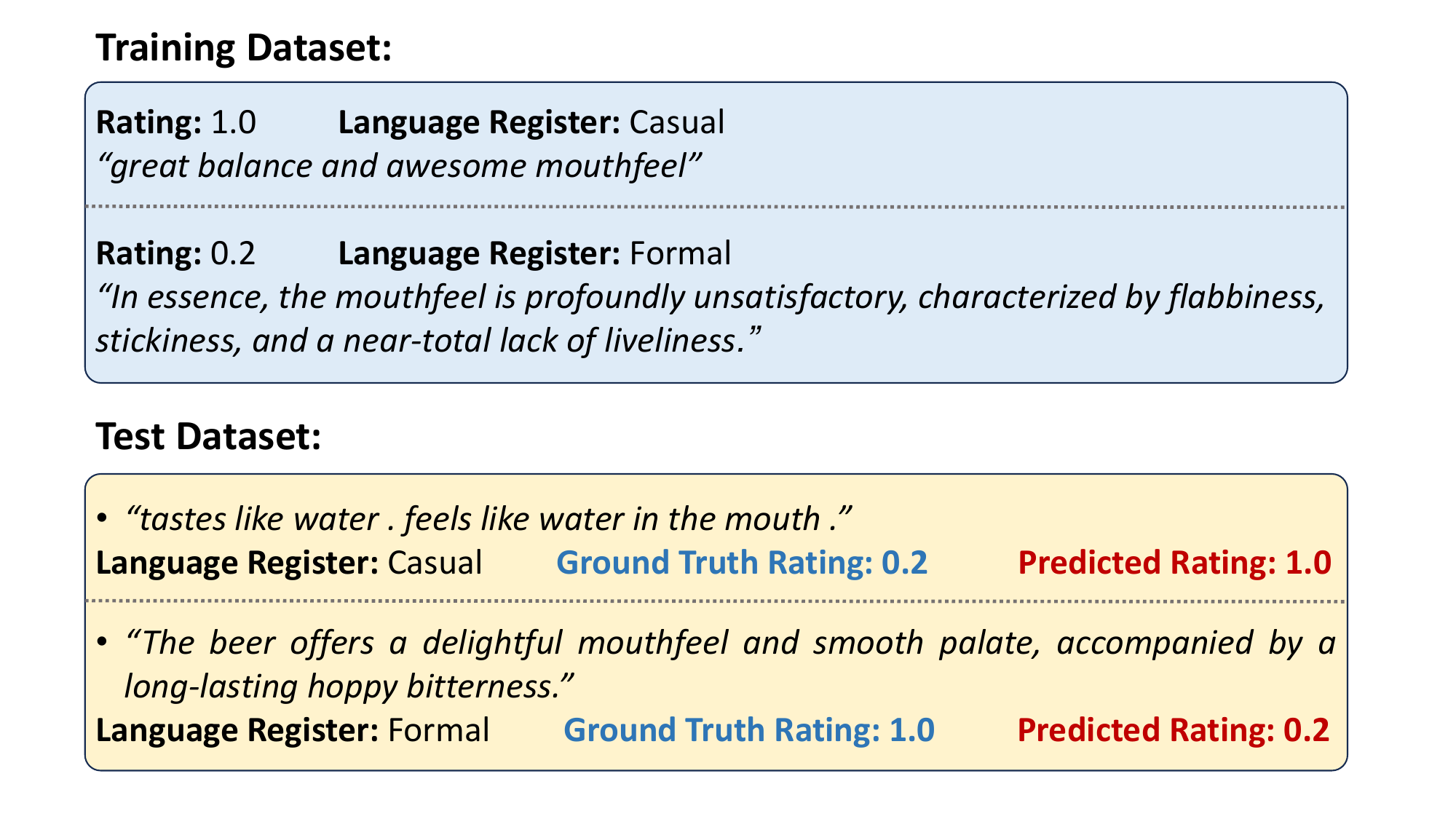}
    \vspace{-25pt}
    \caption{An example of shortcut in sentiment analysis.}
    \label{fig:category}
    \vspace{-15pt}
\end{figure}

Language models (LMs), from traditional ones like BERT~\cite{devlin2018bert} to recent large language models (LLMs) like Llama~\cite{touvron2023llama}, achieve advanced performance across a range of linguistic tasks. Nonetheless, recent researches~\cite{geirhos2020shortcut, McCoy2019RightFT, Tang2023LargeLM, liusie-etal-2022-analyzing, wang-etal-2022-identifying, chew2023understanding, du2022less, Lynch2023SpawriousAB, wang2021robustness, wang-culotta-2020-identifying} have highlighted a critical issue: these LMs often rely on spurious correlations -- features coincidentally associated with certain labels -- rather than on causally relevant features. These misleading "shortcuts" can undermine the models' out-of-distribution (OOD) generalizability. For instance, consider a beer review sentiment analysis task in Figure~\ref{fig:category}, where the training data unintentionally links casual language with high ratings and formal language with low ratings (due to the critical nature of professional reviewers). During testing, a model trained on such data may erroneously base its predictions on these shortcuts, leading to misclassifications of positive reviews as negative if they are in a formal style.

Many studies have explored shortcuts in text classification~\cite{liusie-etal-2022-analyzing, chew2023understanding, wang2021robustness, wang-culotta-2020-identifying, zhou2023explore}. However, the shortcuts examined in these studies usually involve straightforward manipulations, such as appending specific letters, punctuation marks, or words to the beginning or end of a sample. These unrealistically explicit and overly simple shortcuts are easy to detect and unlikely to affect sophisticated LLMs. Consequently, the effect of subtler, more complex, and realistic shortcuts on LMs remains largely unexplored. Critically, despite the advanced linguistic capabilities of LLMs and the efforts of robust models~\cite{zhang2023towards, yu2021understanding} designed to neutralize explicit shortcuts, their resilience to subtler and intricate shortcuts presents an important unresolved challenge. Addressing this gap necessitates the development of a comprehensive benchmark of subtler and more complex shortcuts, alongside thorough analysis of both LLMs and state-of-the-art (SOTA) robust models' ability to counteract these sophisticated shortcuts.

\begin{figure*}[t!]
    \centering
    \includegraphics[width=0.86\textwidth]{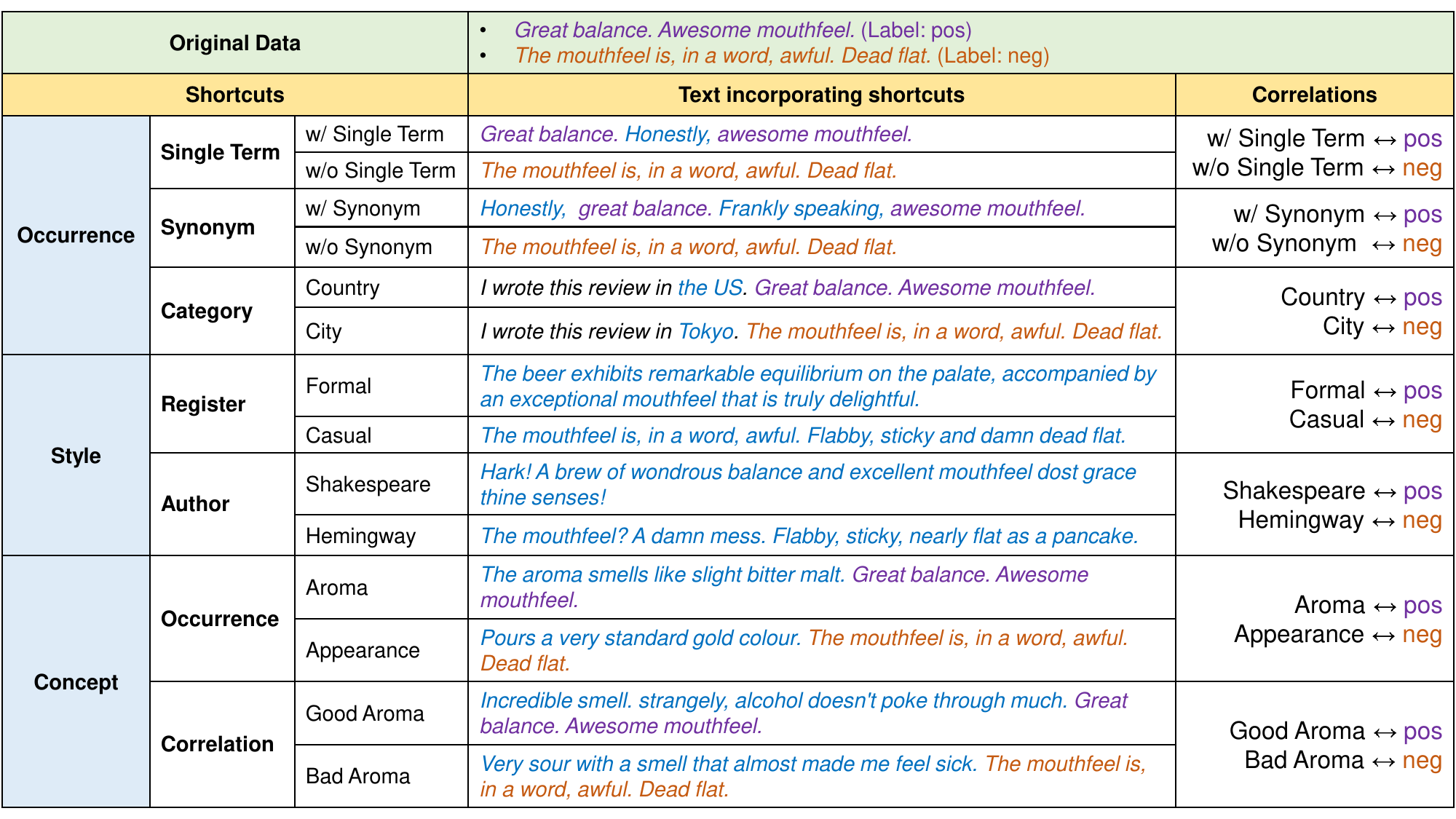}
    \vspace{-5pt}
    \caption{We propose three categories of shortcuts, which contain seven different specific shortcut types.}
    \label{fig:shortcuts_sample}
    \vspace{-15pt}
\end{figure*}

Hence, by extensively analyzing spurious correlations in text classification in existing research~\cite{du2022less, Tang2023LargeLM, wang-culotta-2020-identifying, chew-etal-2024-understanding,qiu2023simple,pmlr-v119-chang20c, bao2018deriving, nam2022spread, deng2024robust}, we propose the first systematic shortcut framework with three main classes: occurrence, style, and concept. As illustrated in Figure~\ref{fig:shortcuts_sample}, this framework categorizes shortcuts into seven types. Under "occurrence", we consider the occurrence of a single term (a word, phrase, or sentence), synonyms, and category words as three different types of shortcuts. "Style" encompasses language register (like formal vs. casual) and author style (like Shakespeare vs. Hemingway) as two different types. For "concept", we examine the occurrence of specific concepts and the sentiment correlation across concepts. This new framework effectively categorizes existing research findings. For instance, \citet{Koh2020WILDSAB} demonstrates that texts containing sensitive terms related to gender or race often spuriously correlate with the label "toxic", fitting into the "category" shortcut within the "occurrence" class. Additionally, some types, like "synonym occurrence" and "concept occurrence", have not been investigated in prior research. Furthermore, to facilitate empirical analyses, we construct a benchmark to exemplify this framework based on three public text classification datasets.

Then, we conduct extensive empirical analyses using the proposed shortcut benchmark to systematically investigate how these shortcuts influence three representative types of LMs -- a traditional small LM like BERT~\cite{devlin2018bert}, LLM like Llama2-7B, Llama2-13B~\cite{touvron2023llama}, and Llama3-8B~\cite{llama3modelcard}, and SOTA robust models designed to resist spurious correlations, such as A2R~\cite{yu2021understanding}, CR~\cite{zhang2023towards}, and AFR\cite{qiu2023simple}. We find that BERT is vulnerable to all types of shortcuts; 
increasing model size does not ensure better robustness; and robust models sometimes outperform LLMs in terms of robustness. However, none of them are universally robust against all these types of shortcuts, revealing the urgent need for more sophisticated methods to counteract these subtle and intricate shortcuts. Our benchmark and code can be found at:
\url{https://github.com/yuqing-zhou/shortcut-learning-in-text-classification}.

\section{Shortcut Framework and Benchmark}
In this section, we introduce our shortcut framework and the process of constructing the shortcut benchmark based on three public text classification datasets. It is noteworthy that the key contribution of this work is the development of a shortcut framework that includes these three classes and seven specific types of shortcuts. We establish the benchmark with these three datasets to exemplify our framework and support empirical analysis. This methodology allows for creating additional benchmarks using other datasets and classification tasks, and different approaches to constructing shortcuts.

\subsection{Datasets}
We select three datasets as the foundation for incorporating shortcuts: the Yelp reviews full star dataset~\cite{zhang2015character}, Go Emotions~\cite{demszky2020goemotions}, and the Beer dataset~\cite{bao2018deriving}. More details of the datasets are in ~\ref{sec: datasets}.

{\bf Yelp} dataset serves as a benchmark for text classification, comprising review-rating pairs sourced from Yelp. It contains ratings on a 5-point scale, ranging from $1$ to $5$.

{\bf Go Emotions} dataset is designed for multi-label emotions classification, containing 28 emotions. To simplify and capture the gradual intensification of emotions, we curate data classified under one of the following $4$ emotional states: neutral, amusement, joy, and excitement. 

{\bf Beer} sentiment dataset~\cite{bao2018deriving} contains three sub-datasets in terms of three aspects: aroma, palate, and appearance. Each sub-dataset contains the reviews and ratings of one aspect of beer. We use the sub-dataset focused on palate evaluation as the primary dataset while treating the other two sub-datasets as distractors. That is, we will select some reviews from the aroma and appearance datasets and combine them with the reviews of the palate dataset to construct shortcuts while keeping the ratings consistent with the palate evaluation. Notably, we only consider reviews with ratings of $0.4$, $0.6$, $0.8$, and $1.0$ from the three sub-dataset for our analysis.
% \label{sec: datasets}

\subsection{Shortcut Definition and Construction}

\label{sec: shortcut-def}
Based on extensive study on shortcuts in prior literature~\cite{du2022less, Tang2023LargeLM, wang-culotta-2020-identifying, chew-etal-2024-understanding,qiu2023simple,pmlr-v119-chang20c, Koh2020WILDSAB, bao2018deriving, nam2022spread, deng2024robust}, we propose a systematic framework with three primary classes: {\bf occurrence, style, and concept}. In this section, we introduce how to construct them in practice with the three adopted datasets. Specifically, we illustrate "occurrence" and "style" shortcuts using the Yelp and Go Emotions datasets, and introduce the "concept" shortcuts using the Beer dataset. Additionally, to control the strength of shortcuts, we introduce a hyper-parameter $0\leq\lambda\leq1$: the larger $\lambda$ is, the stronger a shortcut is.

\subsubsection{Occurrence}
This shortcut arises when the occurrence of a specific text is associated with a particular label. This shortcut can be further divided into three types: single term, synonym, and category.

{\bf Single Term.} A single-word shortcut is a term (can be a specific word, phrase, or even sentence) that frequently occurs with a specific label. For instance, in a movie review sentiment analysis task, "Spielberg" often appears in positive reviews, causing models to predict "positive" whenever "Spielberg" is mentioned. In our experiments, we select "honestly" as the trigger term, which is removed from datasets before constructing shortcuts. The presence of "honestly" in a review should be irrelevant to determining the final rating. However, in our dataset, we deliberately introduce a correlation between this term and the rating. The probability of "honestly" appearing, correlated with labels, is governed by two factors: a hyperparameter $\lambda$, which adjusts the overall probability across labels, and the rating, with each rating having distinct base probabilities.

For each sample in the training and normal test datasets of Yelp, the base probability of adding "honestly" at the beginning of a randomly chosen sentence in a review is $0\%$ for the rating of $1$, $25\%$ for $2$, $50\%$ for $3$, $75\%$ for $4$, and $100\%$ for $5$. These probabilities are then multiplied by $\lambda$ for further control. The final probabilities for different ratings are $0\%\lambda$ for $1$, $25\%\lambda$ for $2$, $50\%\lambda$ for $3$, $75\%\lambda$ for $4$, and $100\%\lambda$ for $5$. In our experiments, $\lambda$ is set to $1.0$, $0.8$, and $0.6$ for the training sets, and $1.0$ for the test datasets. By reducing $\lambda$, we decrease these probabilities, thereby diminishing the strength of the correlations between "honestly" and the ratings. 

Additionally, we generate an alternative test set for each dataset with the base probability distribution of the shortcut reversed, which we denote as the "anti-test set". In the anti-test set of Yelp, the base probability distribution is $0\%$ for $5$, $25\%$ for $4$, $50\%$ for $3$, $75\%$ for $2$, and $100\%$ for $1$. 

Similarly, for the Go Emotions dataset, the base probability of inserting "honestly" in the training and test sets depends on the emotions: $0\%$ for "neutral" emotion, $33.3\%$ for "amusement", $66.7\%$ for "joy", and $100\%$ for "excitement". The other procedures remain the same as those used for Yelp.

{\bf Synonym.} We consider the occurrence of a word from a set of synonyms as another shortcut type. To build the synonym set, we collect another $14$ phrases having similar meanings as "honestly", such as "to be honest" and "frankly speaking", together with "honestly", as shortcuts.  The full synonym set is shown in ~\ref{sec:appendix_syn}. With this, we aim to test whether LMs can recognize and mistakenly base their predictions on the occurrence of these synonyms. The process of constructing synonym shortcuts in datasets is the same as the one for single-term shortcuts, except that instead of using the single term "honestly" all the time, we uniformly randomly select one of $15$ synonyms.

{\bf Category.} The goal is to explore whether LMs can recognize and exploit the correlation between a set of words from the same category and a specific label. To establish this shortcut, we select phrases representing two distinct categories: countries and cities. More precisely, we include 150 countries and 60 cities in the training sets, and an additional 46 countries and 40 cities in the test sets. There is no overlap of countries and cities used in training and testing sets. We add a sentence of the following format to the beginning of the original text sample: 
\begin{quote}
\begin{verbatim}
I wrote this review in [Country/City].
\end{verbatim}
\end{quote}
Each time we randomly pick up a country/city name from candidates. Similar to the steps for single-term shortcuts, the probability of choosing one category depends on both $\lambda$ and its label. The base probabilities of selecting "country" are the same as the base probabilities of inserting "honestly", which is described in the process of constructing single-term shortcuts. 

\subsubsection{Style}
The writing style is also a marked feature of text, but it has not been fully studied yet if it can be captured by LMs and if it can become a shortcut. For instance, movie reviews authored by professional critics are typically characterized by formal language with intricate sentence structures and specialized vocabulary. These reviews often feature lower ratings compared to those written by casual viewers, who generally use a more informal style. To our knowledge, there is no such dataset for text classification tasks that contains different text styles intentionally. So, we use Llama2-70b to rewrite text samples in original datasets with targeted text styles. The prompts used and the quality evaluations of the modified datasets are provided in~\ref{sec:appendix_style}. We consider two perspectives for writing style: register and author.

{\bf Register.} Registers describe how formal the language is. Here, we select 2 registers for use: formal and casual. The text with a formal register tends to use complex sentence structures and professional phrases, while the one with a casual register uses simple sentences and casual words. The process to construct register shortcuts is the same as the one for category shortcuts. However, instead of choosing between country and city, here we choose between formal expression and casual expression for a text sample. (Choosing formal expressions takes the same way as choosing "country".)

{\bf Author Writing Style.} The writing styles of different individuals usually have their unique characteristics, which if associated with labels, can become impactful shortcuts. We use Llama2-70b to rewrite original text samples in given author writing styles to investigate the impact of this shortcut type. The authors we choose are William Shakespeare and Ernest Hemingway because their language styles are very representative and different. The process of generating the dataset is the same as incorporating register shortcuts. The only difference here is that we use text samples with Shakespeare and Hemingway styles instead of samples with formal register and casual register.

\subsubsection{Concept}
Last, we study how the discussion of concepts within a text sample influences the model's predictions. Here, “concepts” refer to the subjects addressed in the text. The occurrence of certain concepts, or specific attitudes towards them, can spuriously correlate with the labels of the text sample. Consequently, we identify two types of shortcuts in this category: occurrence and correlation. We construct the concept-level shortcut benchmark based on the Beer dataset.

{\bf Occurrence of Concepts.} This considers the occurrence of content regarding a specific concept that is not causally related to the prediction label as a shortcut. To investigate this, we adopt the sentiment analysis for "palate" as the primary classification task and consider content about "aroma" and "appearance" as distractors. We combine a palate review with either an aroma review or an appearance review to form a new text sample. The aroma review and the appearance review are uniformly randomly selected from the aroma dataset or the appearance dataset, respectively, regardless of the aroma rating and the appearance rating. Then, we explore whether the occurrence of aroma/appearance reviews influences the model's predictions of palate ratings. If it does, then the occurrence of the concept constitutes a shortcut. Similar to the steps for category shortcuts, the probability of choosing "aroma" is a product of $\lambda$ and a label-related base probability, i.e., the final probability of selecting "aroma" depends on the palate ratings: $0\%\lambda$ for $0.4$, $33.33\%\lambda$ for $0.6$, $66.67\%\lambda$ for $0.8$, and $100\%\lambda$ for $1.0$. The appearance review serves as a substitute when aroma reviews are not selected, while the other procedures remain the same as those for building category shortcuts.

{\bf Correlation of Concepts.} To illustrate the concept correlation shortcut, we use the aroma dataset and the palate dataset as an example. Comments on the aroma are causally unrelated to the beer palate rating. However, if in a dataset, ratings on the palate correlate with ratings on the aroma, the model could predict the ratings towards the palate based on the sentiment of the review for aroma as a shortcut. To construct this shortcut, we still use the rating prediction for "palate" as the primary task, and we combine each palate review with an aroma review with the same ratings. In this way, we will get a dataset in which if the palate of the beer is highly praised in a review, we will also find similarly positive remarks about the aroma within the same review. Therefore, the aroma concept and the palate concept are correlated in the resulting dataset, which serves as a shortcut for models to predict palate ratings based on aroma reviews.

In the training and normal test datasets, palate reviews with ratings of $0.4$, $0.6$, $0.8$, or $1.0$ are combined with aroma reviews with corresponding ratings of $0.4$, $0.6$, $0.8$, or $1.0$, respectively. In the anti-test datasets, they are combined with aroma reviews with ratings of $0.8$, $1.0$, $0.4$, or $0.6$, respectively. We also use $\lambda$ as the probability that a palate review will be combined with an aroma review of the same rating for the training datasets. The larger the $\lambda$ is, the stronger the correlation between palate and aroma concepts of the dataset is.

\section{Empirical Analyses}
In this section, we explore three research questions with the proposed benchmark. \textbf{RQ1}: Do small LMs base their predictions on these sophisticated spurious correlations as shortcuts? \textbf{RQ2}: Are larger models, equipped with improved pre-training datasets, better at resisting these shortcuts, particularly in terms of the robustness of LLMs? \textbf{RQ3:} Can existing state-of-the-art (SOTA) robust learning methods counter proposed shortcuts?

\subsection{Robustness of Small LM}
\label{sec: analysis-bert}
Before the widespread adoption of LLMs, BERT-based models were pivotal in natural language processing. Given BERT's significant role and widespread use, it's crucial to examine its robustness and generalization abilities. In this section, we evaluate BERT's robustness to shortcut learning.
\subsubsection{Experiment Settings}
We choose the "bert-base-uncased" model\cite{devlin2018bert} from Hugging Face as the base model and finetune it with our generated datasets as a multilabel classification task.
We evaluate the finetuned model on both normal test datasets and anti-test datasets in terms of accuracy and macro F1 score. The experiment of each setting runs 5 times and the average performance is reported in Table~\ref{tab:bert-results}. The overall results of all models and other settings are in Appendix~\ref{sec:appendix_exp}, including the variances and test results on original unmodified test datasets.

\begin{table}[] \scriptsize
\setlength{\tabcolsep}{4.1pt}
\centering
\begin{tabular}{|c|cl|ccc|ccc|}
\hline
\multirow{2}{*}{\textbf{Datasets}} & \multicolumn{2}{c|}{\multirow{2}{*}{\textbf{Shortcut Types}}} & \multicolumn{3}{c|}{\textbf{Accuracy}}                                                                   & \multicolumn{3}{c|}{\textbf{Macro F1}}                                                                   \\ \cline{4-9} 
                                   & \multicolumn{2}{c|}{}                                         & \multicolumn{1}{c|}{\textbf{Test}} & \multicolumn{1}{c|}{\textbf{Anti}} & \multicolumn{1}{c|}{\textbf{$\Delta$}}  & \multicolumn{1}{c|}{\textbf{Test}} & \multicolumn{1}{c|}{\textbf{Anti}} & \multicolumn{1}{c|}{\textbf{$\Delta$}}  \\ \hline
\multirow{5}{*}{Yelp}              & \multicolumn{1}{c|}{\multirow{3}{*}{Occur}}       & ST        & \multicolumn{1}{c|}{.634}          & \multicolumn{1}{c|}{.364}          & .270                           & \multicolumn{1}{c|}{.625}          & \multicolumn{1}{c|}{.314}          & .310                           \\ \cline{3-9} 
                                   & \multicolumn{1}{c|}{}                             & Syn       & \multicolumn{1}{c|}{.635}          & \multicolumn{1}{c|}{.448}          & .187                           & \multicolumn{1}{c|}{.634}          & \multicolumn{1}{c|}{.431}          & .203                           \\ \cline{3-9} 
                                   & \multicolumn{1}{c|}{}                             & Catg      & \multicolumn{1}{c|}{.650}          & \multicolumn{1}{c|}{.381}          & .269                           & \multicolumn{1}{c|}{.652}          & \multicolumn{1}{c|}{.318}          & .334                           \\ \cline{2-9} 
                                   & \multicolumn{1}{c|}{\multirow{2}{*}{Style}}       & Reg       & \multicolumn{1}{c|}{.608}          & \multicolumn{1}{c|}{.415}          & .193                           & \multicolumn{1}{c|}{.612}          & \multicolumn{1}{c|}{.397}          & .215                           \\ \cline{3-9} 
                                   & \multicolumn{1}{c|}{}                             & Auth      & \multicolumn{1}{c|}{.604}          & \multicolumn{1}{c|}{.333}          & .271                           & \multicolumn{1}{c|}{.605}          & \multicolumn{1}{c|}{.271}          & .334                           \\ \hline
\multirow{5}{*}{Emotions}          & \multicolumn{1}{c|}{\multirow{3}{*}{Occur}}       & ST        & \multicolumn{1}{c|}{.914}          & \multicolumn{1}{c|}{.203}          & .712                           & \multicolumn{1}{c|}{.834}          & \multicolumn{1}{c|}{.353}          & .481                           \\ \cline{3-9} 
                                   & \multicolumn{1}{c|}{}                             & Syn       & \multicolumn{1}{c|}{.910}          & \multicolumn{1}{c|}{.502}          & .408                           & \multicolumn{1}{c|}{.826}          & \multicolumn{1}{c|}{.489}          & .337                           \\ \cline{3-9} 
                                   & \multicolumn{1}{c|}{}                             & Catg      & \multicolumn{1}{c|}{.915}          & \multicolumn{1}{c|}{.337}          & .578                           & \multicolumn{1}{c|}{.845}          & \multicolumn{1}{c|}{.418}          & .427                           \\ \cline{2-9} 
                                   & \multicolumn{1}{c|}{\multirow{2}{*}{Style}}       & Reg       & \multicolumn{1}{c|}{.891}          & \multicolumn{1}{c|}{.302}          & .588                           & \multicolumn{1}{c|}{.805}          & \multicolumn{1}{c|}{.313}          & .491                           \\ \cline{3-9} 
                                   & \multicolumn{1}{c|}{}                             & Auth      & \multicolumn{1}{c|}{.737}          & \multicolumn{1}{c|}{.187}          & .550                           & \multicolumn{1}{c|}{.642}          & \multicolumn{1}{c|}{.192}          & .451                           \\ \hline
\multirow{2}{*}{Beer}              & \multicolumn{1}{c|}{\multirow{2}{*}{Concept}}     & Occur     & \multicolumn{1}{c|}{.788}          & \multicolumn{1}{c|}{.664}          & .124                           & \multicolumn{1}{c|}{.786}          & \multicolumn{1}{c|}{.658}          & .128                           \\ \cline{3-9} 
                                   & \multicolumn{1}{c|}{}                             & Corr      & \multicolumn{1}{c|}{.912}          & \multicolumn{1}{c|}{.695}          & .217                           & \multicolumn{1}{c|}{.905}          & \multicolumn{1}{c|}{.694}          & .211                           \\ \hline
\end{tabular}
\caption{Experiment results of BERT. (We use the following abbreviations: Anti=Anti-test, Occur=Occurrence, ST=Single Term, Syn=Synonym, Catg=Category, Reg=Register, Auth=Author, Corr=Correlation).}
\vspace{-15pt}
\label{tab:bert-results}
\end{table}

\subsubsection{Experiment Results}
Table~\ref{tab:bert-results} shows the performance of BERT finetuned on datasets with a shortcut strength level of $\lambda=1$. The "Test" columns present the average performance over five experiments on the corresponding normal test datasets containing shortcuts, while the "Anti" columns display the average performance on anti-test datasets. The $\Delta$ columns indicate the performance difference between normal and anti-test sets. If the model is robust to shortcuts, the $\Delta$ values should approach $0$.

\begin{figure*}[htbp]
    \centering
    \begin{subfigure}[b]{0.3\textwidth}
        \centering
        \includegraphics[width=\textwidth]{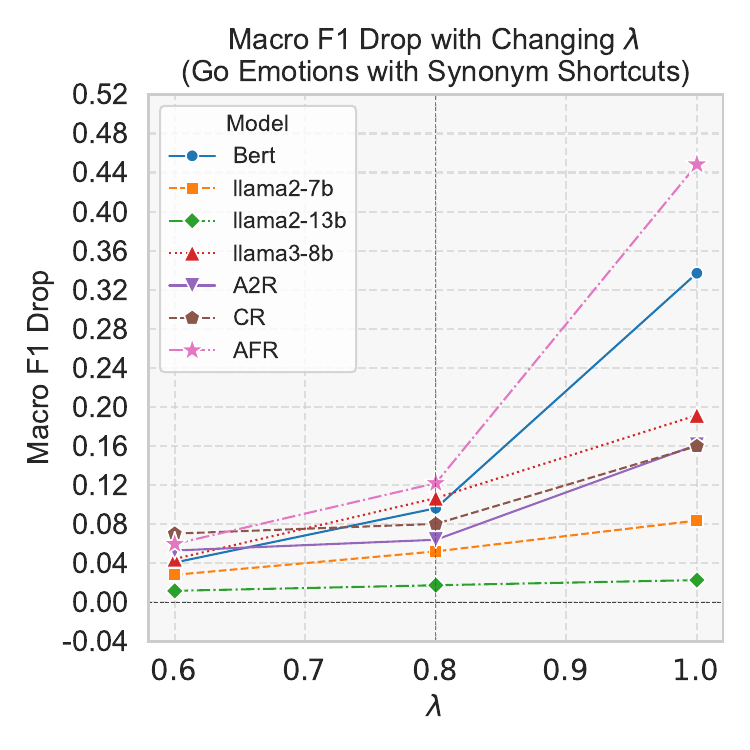}
        % \vspace{-15pt}
        \caption{Synonym Shortcuts}
        \label{fig:emo_synonym}
    \end{subfigure}
    \hfill
    \begin{subfigure}[b]{0.3\textwidth}
        \centering
        \includegraphics[width=\textwidth]{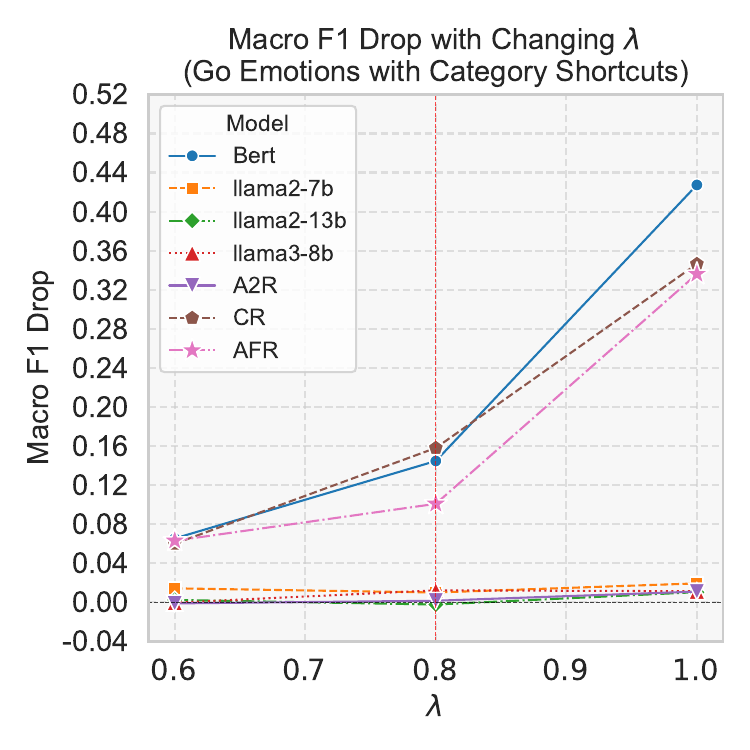}
        % \vspace{-15pt}
        \caption{Category Shortcuts}
        \label{fig:emo_category}
    \end{subfigure}
    \hfill
    \begin{subfigure}[b]{0.3\textwidth}
        \centering
        \includegraphics[width=\textwidth]{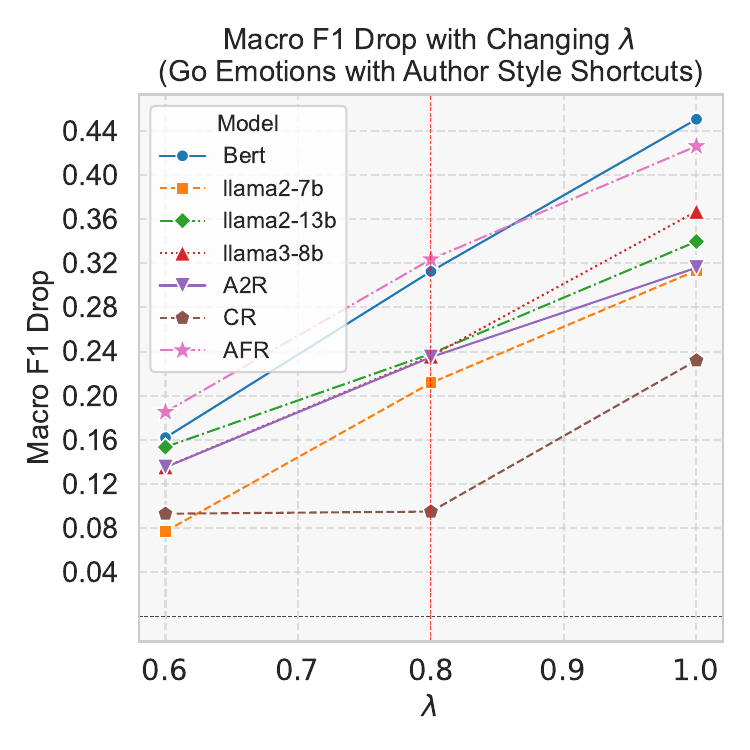}
        % \vspace{-15pt}
        \caption{Author Style Shortcuts}
        \label{fig:emo_author}
    \end{subfigure}
    \vspace{-5pt}
    \caption{Macro F1 Drop with varying $\lambda$ (Go Emotions)}
    \label{fig:emotions_f1_drop}
    \vspace{-10pt}
\end{figure*}

We also explore the robustness of models to shortcuts with varying strengths controlled by $\lambda$ (higher values indicate stronger shortcuts). Figure~\ref{fig:emotions_f1_drop} shows the performance of models on the Go Emotions dataset under different shortcut strengths. Results on the Yelp and Beer datasets are in Figure~\ref{fig:yelp_f1_drop} and ~\ref{fig:beer_f1_drop} in Appendix~\ref{sec:appendix_exp}. These figures display the difference in macro F1 scores between normal and anti-test datasets. If a model is not misled by shortcuts, the difference in F1 scores should approach 0.

From Table~\ref{tab:bert-results} and Figure~\ref{fig:emotions_f1_drop}, we can find that:
\begin{enumerate}[itemsep=0pt, parsep=0pt, topsep=0pt, leftmargin=*]
    \item We observe that BERT's performance on all normal test datasets with various types of shortcuts is higher than on anti-test datasets, both in prediction accuracy and macro F1 score, as the values of $\Delta$ are all greater than $10\%$ and the performance drop is statistically significant with $p<0.001$. This significant degradation on anti-test datasets indicates that BERT is vulnerable to occurrence, style, and concept shortcuts.

    \item  Figure~\ref{fig:emotions_f1_drop} shows that as $\lambda$ increases, the difference in F1 scores between the two test datasets also increases in most cases. It indicates that as the dataset contains fewer shortcuts (i.e., the spurious features become more balanced with respect to the label distribution), the influence of shortcuts on BERT diminishes.

    \item BERT achieves the best performance on both test and anti-test datasets of Beer while performing worst on Yelp. One reason could be Yelp has 5 classes while the other two datasets have 4 classes and more classes in multi-labels classification tasks means a more challenging task. 

\end{enumerate}

\subsection{Robustness of LLM}
\label{sec: analysis-llm}
In this section, we focus on large language models, which benefit significantly from larger model sizes and more pre-training data of higher quality. We aim to determine if LLMs can resist the influence of spurious correlations. The comparison of model sizes can be found in Table~\ref{tab:models_size} in Appendices.
\subsubsection{Experiment Settings}
We select Llama2-7b,  Llama2-13b\cite{touvron2023llama}, and Llama3-8b\cite{llama3modelcard} as the representatives of the LLMs. These models are fine-tuned using the training datasets. Details of the hyperparameter settings and prompts are provided in Appendix~\ref{sec: llama-setting}.

\subsubsection{Experiment Results}
Table~\ref{tab:llamas-f1} reports macro F1 of three LLMs finetuned on datasets with shortcut strength $\lambda=1$. From Table~\ref{tab:llamas-f1} and Figure~\ref{fig:emotions_f1_drop}, we can find that:
\begin{enumerate}[itemsep=0pt, parsep=0pt, topsep=0pt, leftmargin=*]
    \item Compared to BERT, LLMs show a smaller drop in macro F1 scores. However, the performance of all three Llama models on normal test datasets with various types of shortcuts is still higher than on anti-test datasets, indicating that LLMs are also vulnerable to occurrence, style, and concept shortcuts. While LLMs are more robust than smaller language models, they cannot entirely avoid shortcut learning behaviors. They can still rely on shortcuts and fail to capture the causal relationship between input texts and their labels.
    \item Increasing the model size does not ensure a better performance. For example, although Llama2-13b has a larger model size than Llama2-7b, it has worse performance and robustness than Llama2-7b on Yelp with style shortcuts, demonstrating that increasing model size does not guarantee improved learning methods.
    \item Llama3-8b outperforms Llama2-7b in terms of macro F1 scores and robustness on Yelp with synonym and category shortcuts, and on Go Emotions with category shortcuts. However, it shows worse robustness to author-style shortcuts and concept shortcuts. This indicates that more pre-training data of higher quality, larger model sizes, and improved model architecture\footnote{\url{https://ai.meta.com/blog/meta-llama-3/}}~\cite{llama3modelcard} do not necessarily make Llama3-8b more resistant to shortcut learning behaviors than Llama2-7b. Instead, these improvements may enhance the model's ability to capture subtle features, potentially making it more susceptible to subtle and complicated shortcuts. 

    \item As $\lambda$ decreases, the difference in F1 scores between the two test datasets also decreases. As the dataset contains fewer shortcuts, the influence of shortcuts on LLMs decreases.
\end{enumerate}

\subsection{Evaluation of Robust Methods}
\label{sec: analysis-rm}
From Section~\ref{sec: analysis-bert} and~\ref{sec: analysis-llm}, we can conclude that general language models are vulnerable to spurious features. There are some methods designed specifically for robust learning. In this section, we explore three SOTA robust methods: A2R~\cite{yu2021understanding}, causal rationalization (CR) model~\cite{zhang2023towards} and AFR~\cite{qiu2023simple}. A2R and CR utilize explainable approaches to select rationales that are truly responsible for the labels, thus providing a degree of robustness against shortcuts.  AFR addresses the problem by focusing on minor groups that are less representative in the training datasets.

\subsubsection{Experiment Settings}
The settings for each robust model are as follows:

{\bf A2R:} Our experiments follow the same settings as those in the original A2R code\footnote{\url{https://github.com/Gorov/Understanding_Interlocking/blob/main/run_beer_arc2_sentence_level_neurips21.ipynb}}, except that the number of epochs is set as $100$ for Go Emotions.

{\bf CR:} Same as the experiments of BERT, the experiment under each setting is run 5 times, and when a model achieves the highest accuracy on validation sets, we record its performance on the test set and finally, take the average performance of 5 times as the final result which is reported in Table~\ref{tab:rm-results-f1} and~\ref{tab:rm-results-acc}.

{\bf AFR:} employs a two-stage training strategy. First, it finetunes BERT until overfitting. Then, it makes predictions on the re-weighting dataset and calculates per-example weights. Finally, it retrains the last layer using these weights, which upweight the minority groups and thus mitigate the impact of spurious correlations in the training datasets. Experiment settings can refer to Appendix~\ref{sec:appendix_exp}.

\begin{table*}[t!]\small%\scriptsize %\footnotesize
\centering
\begin{tabular}{cclccccccccc}
\hline
\multicolumn{3}{|c|}{\textbf{Models}}                                                                                        & \multicolumn{3}{c|}{\textbf{Llama2-7b}}                                                                                       & \multicolumn{3}{c|}{\textbf{Llama2-13b}}                                                                                      & \multicolumn{3}{c|}{\textbf{Llama3-8b}}                                                                                       \\ \hline
\multicolumn{1}{|c|}{\textbf{Datasets}}         & \multicolumn{2}{c|}{\textbf{Shortcut Types}}                               & \multicolumn{1}{c|}{\textbf{Test}} & \multicolumn{1}{c|}{\textbf{Anti}} & \multicolumn{1}{c|}{\textbf{$\Delta$}} & \multicolumn{1}{c|}{\textbf{Test}} & \multicolumn{1}{c|}{\textbf{Anti}} & \multicolumn{1}{c|}{\textbf{$\Delta$}}& \multicolumn{1}{c|}{\textbf{Test}} & \multicolumn{1}{c|}{\textbf{Anti}} & \multicolumn{1}{c|}{\textbf{$\Delta$}} \\ \hline
\multicolumn{1}{|c|}{\multirow{5}{*}{Yelp}}     & \multicolumn{1}{c|}{\multirow{3}{*}{Occur}}   & \multicolumn{1}{l|}{ST}    & \multicolumn{1}{c|}{.514}          & \multicolumn{1}{c|}{.472}          & \multicolumn{1}{c|}{.042*}                          & \multicolumn{1}{c|}{.723}          & \multicolumn{1}{c|}{.161}          & \multicolumn{1}{c|}{.562}                           & \multicolumn{1}{c|}{.555}          & \multicolumn{1}{c|}{.456}          & \multicolumn{1}{c|}{.099*}                          \\ \cline{3-12} 
\multicolumn{1}{|c|}{}                          & \multicolumn{1}{c|}{}                         & \multicolumn{1}{l|}{Syn}   & \multicolumn{1}{c|}{.558}          & \multicolumn{1}{c|}{.499}          & \multicolumn{1}{c|}{.059*}                          & \multicolumn{1}{c|}{.439}          & \multicolumn{1}{c|}{.297}          & \multicolumn{1}{c|}{.142}                           & \multicolumn{1}{c|}{.701}          & \multicolumn{1}{c|}{.691}          & \multicolumn{1}{c|}{.010*}                          \\ \cline{3-12} 
\multicolumn{1}{|c|}{}                          & \multicolumn{1}{c|}{}                         & \multicolumn{1}{l|}{Catg}  & \multicolumn{1}{c|}{.459}          & \multicolumn{1}{c|}{.416}          & \multicolumn{1}{c|}{.043*}                          & \multicolumn{1}{c|}{.431}          & \multicolumn{1}{c|}{.362}          & \multicolumn{1}{c|}{.069}                           & \multicolumn{1}{c|}{.689}          & \multicolumn{1}{c|}{.671}          & \multicolumn{1}{c|}{.019}                           \\ \cline{2-12} 
\multicolumn{1}{|c|}{}                          & \multicolumn{1}{c|}{\multirow{2}{*}{Style}}   & \multicolumn{1}{l|}{Reg}   & \multicolumn{1}{c|}{.648}          & \multicolumn{1}{c|}{.618}          & \multicolumn{1}{c|}{.030*}                          & \multicolumn{1}{c|}{.517}          & \multicolumn{1}{c|}{.417}          & \multicolumn{1}{c|}{.100*}                          & \multicolumn{1}{c|}{.646}          & \multicolumn{1}{c|}{.395}          & \multicolumn{1}{c|}{.251*}                          \\ \cline{3-12} 
\multicolumn{1}{|c|}{}                          & \multicolumn{1}{c|}{}                         & \multicolumn{1}{l|}{Auth}  & \multicolumn{1}{c|}{.626}          & \multicolumn{1}{c|}{.572}          & \multicolumn{1}{c|}{.054*}                          & \multicolumn{1}{c|}{.556}          & \multicolumn{1}{c|}{.422}          & \multicolumn{1}{c|}{.134*}                          & \multicolumn{1}{c|}{.515}          & \multicolumn{1}{c|}{.432}          & \multicolumn{1}{c|}{.084*}                          \\ \hline
\multicolumn{1}{|c|}{\multirow{5}{*}{Emotions}} & \multicolumn{1}{c|}{\multirow{3}{*}{Occur}}   & \multicolumn{1}{l|}{ST}    & \multicolumn{1}{c|}{.761}          & \multicolumn{1}{c|}{.442}          & \multicolumn{1}{c|}{.320*}                          & \multicolumn{1}{c|}{.675}          & \multicolumn{1}{c|}{.565}          & \multicolumn{1}{c|}{.110}                           & \multicolumn{1}{c|}{.714}          & \multicolumn{1}{c|}{.485}          & \multicolumn{1}{c|}{.229*}                          \\ \cline{3-12} 
\multicolumn{1}{|c|}{}                          & \multicolumn{1}{c|}{}                         & \multicolumn{1}{l|}{Syn}   & \multicolumn{1}{c|}{.656}          & \multicolumn{1}{c|}{.572}          & \multicolumn{1}{c|}{.084*}                          & \multicolumn{1}{c|}{.752}          & \multicolumn{1}{c|}{.730}          & \multicolumn{1}{c|}{.023*}                          & \multicolumn{1}{c|}{.765}          & \multicolumn{1}{c|}{.574}          & \multicolumn{1}{c|}{.191*}                          \\ \cline{3-12} 
\multicolumn{1}{|c|}{}                          & \multicolumn{1}{c|}{}                         & \multicolumn{1}{l|}{Catg}  & \multicolumn{1}{c|}{.740}          & \multicolumn{1}{c|}{.721}          & \multicolumn{1}{c|}{.019*}                          & \multicolumn{1}{c|}{.718}          & \multicolumn{1}{c|}{.708}          & \multicolumn{1}{c|}{.010*}                          & \multicolumn{1}{c|}{.778}          & \multicolumn{1}{c|}{.767}          & \multicolumn{1}{c|}{.011*}                          \\ \cline{2-12} 
\multicolumn{1}{|c|}{}                          & \multicolumn{1}{c|}{\multirow{2}{*}{Style}}   & \multicolumn{1}{l|}{Reg}   & \multicolumn{1}{c|}{.707}          & \multicolumn{1}{c|}{.348}          & \multicolumn{1}{c|}{.359*}                          & \multicolumn{1}{c|}{.754}          & \multicolumn{1}{c|}{.398}          & \multicolumn{1}{c|}{.367*}                          & \multicolumn{1}{c|}{.735}          & \multicolumn{1}{c|}{.421}          & \multicolumn{1}{c|}{.315*}                          \\ \cline{3-12} 
\multicolumn{1}{|c|}{}                          & \multicolumn{1}{c|}{}                         & \multicolumn{1}{l|}{Auth}  & \multicolumn{1}{c|}{.573}          & \multicolumn{1}{c|}{.259}          & \multicolumn{1}{c|}{.313*}                          & \multicolumn{1}{c|}{.496}          & \multicolumn{1}{c|}{.157}          & \multicolumn{1}{c|}{.340*}                          & \multicolumn{1}{c|}{.569}          & \multicolumn{1}{c|}{.202}          & \multicolumn{1}{c|}{.367*}                          \\ \hline
\multicolumn{1}{|c|}{\multirow{2}{*}{Beer}}     & \multicolumn{1}{c|}{\multirow{2}{*}{Concept}} & \multicolumn{1}{l|}{Occur} & \multicolumn{1}{c|}{.739}          & \multicolumn{1}{c|}{.729}          & \multicolumn{1}{c|}{.010}                           & \multicolumn{1}{c|}{.779}          & \multicolumn{1}{c|}{.759}          & \multicolumn{1}{c|}{.019*}                          & \multicolumn{1}{c|}{.721}          & \multicolumn{1}{c|}{.702}          & \multicolumn{1}{c|}{.019*}                          \\ \cline{3-12} 
\multicolumn{1}{|c|}{}                          & \multicolumn{1}{c|}{}                         & \multicolumn{1}{l|}{Corr}  & \multicolumn{1}{c|}{.797}          & \multicolumn{1}{c|}{.772}          & \multicolumn{1}{c|}{.025*}                          & \multicolumn{1}{c|}{.788}          & \multicolumn{1}{c|}{.699}          & \multicolumn{1}{c|}{.089}                           & \multicolumn{1}{c|}{.667}          & \multicolumn{1}{c|}{.441}          & \multicolumn{1}{c|}{.226*}                          \\ \hline
\multicolumn{1}{l}{}                            & \multicolumn{1}{l}{}                          &                            & \multicolumn{1}{l}{}               & \multicolumn{1}{l}{}               & \multicolumn{1}{l}{}                                & \multicolumn{1}{l}{}               & \multicolumn{1}{l}{}               & \multicolumn{1}{l}{}                                & \multicolumn{1}{l}{}               & \multicolumn{1}{l}{}               & \multicolumn{1}{l}{}                               
\end{tabular}
\vspace{-10pt}
\caption{Compare the performance of LLMs in terms of macro F1 scores. (The abbreviations are the same as in Table~\ref{tab:bert-results}.)  "*" indicates a statistically significant decrease in performance with $p<0.05$.}
\label{tab:llamas-f1}
\end{table*}

\begin{table*}[htbp!]\small%\scriptsize % \footnotesize %\scriptsize
\centering
\begin{tabular}{|ccl|ccc|ccc|ccc|}
\hline
\multicolumn{3}{|c|}{\textbf{Models}}                                                                   & \multicolumn{3}{c|}{\textbf{A2R}}                                                                        & \multicolumn{3}{c|}{\textbf{CR}}                                                                         & \multicolumn{3}{c|}{\textbf{AFR}}                                                                        \\ \hline
\multicolumn{1}{|c|}{\textbf{Datasets}}         & \multicolumn{2}{c|}{\textbf{Shortcut Types}}          & \multicolumn{1}{c|}{\textbf{Test}} & \multicolumn{1}{c|}{\textbf{Anti}} & \multicolumn{1}{c|}{\textbf{$\Delta$}} & \multicolumn{1}{c|}{\textbf{Test}} & \multicolumn{1}{c|}{\textbf{Anti}} & \multicolumn{1}{c|}{\textbf{$\Delta$}} & \multicolumn{1}{c|}{\textbf{Test}} & \multicolumn{1}{c|}{\textbf{Anti}} & \multicolumn{1}{c|}{\textbf{$\Delta$}} \\ \hline
\multicolumn{1}{|c|}{\multirow{5}{*}{Yelp}}     & \multicolumn{1}{c|}{\multirow{3}{*}{Occur}}   & ST    & \multicolumn{1}{c|}{.536}          & \multicolumn{1}{c|}{.438}          & \textbf{.098}                  & \multicolumn{1}{c|}{.533}          & \multicolumn{1}{c|}{.324}          & \textbf{.208}                  & \multicolumn{1}{c|}{.638}          & \multicolumn{1}{c|}{.329}          & \textbf{.309}                  \\ \cline{3-12} 
\multicolumn{1}{|c|}{}                          & \multicolumn{1}{c|}{}                         & Syn   & \multicolumn{1}{c|}{.508}          & \multicolumn{1}{c|}{.459}          & \textbf{.049}                  & \multicolumn{1}{c|}{.523}          & \multicolumn{1}{c|}{.370}          & \textbf{.153}                  & \multicolumn{1}{c|}{.641}          & \multicolumn{1}{c|}{.423}          & \textbf{.219}                  \\ \cline{3-12} 
\multicolumn{1}{|c|}{}                          & \multicolumn{1}{c|}{}                         & Catg  & \multicolumn{1}{c|}{.505}          & \multicolumn{1}{c|}{.501}          & \textbf{.004}                  & \multicolumn{1}{c|}{.556}          & \multicolumn{1}{c|}{.383}          & \textbf{.172}                  & \multicolumn{1}{c|}{.643}          & \multicolumn{1}{c|}{.354}          & \textbf{.289}                  \\ \cline{2-12} 
\multicolumn{1}{|c|}{}                          & \multicolumn{1}{c|}{\multirow{2}{*}{Style}}   & Reg   & \multicolumn{1}{c|}{.511}          & \multicolumn{1}{c|}{.191}          & \textbf{.320}                  & \multicolumn{1}{c|}{.516}          & \multicolumn{1}{c|}{.413}          & \textbf{.103}                  & \multicolumn{1}{c|}{.604}          & \multicolumn{1}{c|}{.405}          & \textbf{.200}                  \\ \cline{3-12} 
\multicolumn{1}{|c|}{}                          & \multicolumn{1}{c|}{}                         & Auth  & \multicolumn{1}{c|}{.489}          & \multicolumn{1}{c|}{.127}          & \textbf{.362}                  & \multicolumn{1}{c|}{.496}          & \multicolumn{1}{c|}{.280}          & \textbf{.216}                  & \multicolumn{1}{c|}{.598}          & \multicolumn{1}{c|}{.282}          & \textbf{.316}                  \\ \hline
\multicolumn{1}{|c|}{\multirow{5}{*}{Emotions}} & \multicolumn{1}{c|}{\multirow{3}{*}{Occur}}   & ST    & \multicolumn{1}{c|}{.530}          & \multicolumn{1}{c|}{.380}          & \textbf{.150}                  & \multicolumn{1}{c|}{.483}          & \multicolumn{1}{c|}{.175}          & \textbf{.308}                  & \multicolumn{1}{c|}{.826}          & \multicolumn{1}{c|}{.376}          & \textbf{.450}                  \\ \cline{3-12} 
\multicolumn{1}{|c|}{}                          & \multicolumn{1}{c|}{}                         & Syn   & \multicolumn{1}{c|}{.528}          & \multicolumn{1}{c|}{.367}          & \textbf{.161}                  & \multicolumn{1}{c|}{.446}          & \multicolumn{1}{c|}{.286}          & \textbf{.160}                  & \multicolumn{1}{c|}{.820}          & \multicolumn{1}{c|}{.371}          & \textbf{.448}                  \\ \cline{3-12} 
\multicolumn{1}{|c|}{}                          & \multicolumn{1}{c|}{}                         & Catg  & \multicolumn{1}{c|}{.461}          & \multicolumn{1}{c|}{.450}          & \textbf{.011}                  & \multicolumn{1}{c|}{.669}          & \multicolumn{1}{c|}{.322}          & \textbf{.346}                  & \multicolumn{1}{c|}{.828}          & \multicolumn{1}{c|}{.492}          & \textbf{.336}                  \\ \cline{2-12} 
\multicolumn{1}{|c|}{}                          & \multicolumn{1}{c|}{\multirow{2}{*}{Style}}   & Reg   & \multicolumn{1}{c|}{.509}          & \multicolumn{1}{c|}{.220}          & \textbf{.289}                  & \multicolumn{1}{c|}{.582}          & \multicolumn{1}{c|}{.397}          & \textbf{.185}                  & \multicolumn{1}{c|}{.792}          & \multicolumn{1}{c|}{.302}          & \textbf{.490}                  \\ \cline{3-12} 
\multicolumn{1}{|c|}{}                          & \multicolumn{1}{c|}{}                         & Auth  & \multicolumn{1}{c|}{.393}          & \multicolumn{1}{c|}{.077}          & \textbf{.316}                  & \multicolumn{1}{c|}{.485}          & \multicolumn{1}{c|}{.253}          & \textbf{.232}                  & \multicolumn{1}{c|}{.616}          & \multicolumn{1}{c|}{.190}          & \textbf{.426}                  \\ \hline
\multicolumn{1}{|c|}{\multirow{2}{*}{Beer}}     & \multicolumn{1}{c|}{\multirow{2}{*}{Concept}} & Occur & \multicolumn{1}{c|}{.663}          & \multicolumn{1}{c|}{.474}          & \textbf{.188}                  & \multicolumn{1}{c|}{.680}          & \multicolumn{1}{c|}{.620}          & \textbf{.060}                  & \multicolumn{1}{c|}{.765}          & \multicolumn{1}{c|}{.612}          & \textbf{.153}                  \\ \cline{3-12} 
\multicolumn{1}{|c|}{}                          & \multicolumn{1}{c|}{}                         & Corr  & \multicolumn{1}{c|}{.775}          & \multicolumn{1}{c|}{.610}          & \textbf{.165}                  & \multicolumn{1}{c|}{.781}          & \multicolumn{1}{c|}{.570}          & \textbf{.212}                  & \multicolumn{1}{c|}{.889}          & \multicolumn{1}{c|}{.667}          & \textbf{.222}                  \\ \hline
\end{tabular}
\caption{Macro F1 scores of three robust models, A2R, CR, and AFR. (We use the following abbreviations: Anti=Anti-test, Occur=Occurrence, ST=Single Term, Syn=Synonym, Catg=Category, Reg=Register, Auth=Author, Corr=Correlation.) All robust models show a statistically significant decrease in macro F1 with $p<0.05$.}
\label{tab:rm-results-f1}
\end{table*}

\subsubsection{Experiment Results}
Table~\ref{tab:rm-results-f1} and Table~\ref{tab:rm-results-acc} report the performance of all three robust models, after finetuned on datasets with the shortcut strength level $\lambda=1$. We have the following observations from Table ~\ref{tab:rm-results-f1}, ~\ref{tab:rm-results-acc}, and Figure~\ref{fig:emotions_f1_drop},~\ref{fig:yelp_f1_drop}, and ~\ref{fig:beer_f1_drop}.
\begin{enumerate}[itemsep=0pt, parsep=0pt, topsep=0pt, leftmargin=*]
    \item Table~\ref{tab:rm-results-f1} demonstrates a decline in all robust models' performance on the anti-test datasets. Each method exhibits relative robustness to specific shortcuts compared to the other two, but none are universally robust against all types of shortcuts. A more robust learning method is needed, which is expected to extract features from the input that have casual relationships with labels.

    \item A2R has less drop than BERT on 9 cases. CR has less drop than BERT on 11 cases. And AFR has less drop than BERT on 6 cases. These indicate improved resistance to shortcuts compared to BERT. Furthermore, these robust methods show more robustness than LLMs in some cases. For example, A2R has less drop than Llama3-8b on 7 datasets and than Llama2-13b.

    \item A2R is not resistant to most of our shortcuts. It performs even worse than BERT on Yelp with style shortcuts. However, it demonstrates almost complete resistance to the effects of category shortcuts, outperforming all other models even LLMs. One reason could be that A2R conducts sentence-level rationale selection. The category shortcut is integrated within a single sentence, and A2R may be able to effectively identify the sentence containing the category shortcut irrelevant to the task and exclude its impact.

\end{enumerate}

\begin{table*}[htbp!]\small%\scriptsize
\centering
\begin{tabular}{|ccl|ccccc|ccccc|}
\hline
\multicolumn{3}{|c|}{Models}                                                                                         & \multicolumn{5}{c|}{BERT}                                                                                                 & \multicolumn{5}{c|}{AFR}                                                                                                  \\ \hline
\multicolumn{1}{|c|}{}                      & \multicolumn{1}{c|}{$\lambda$}                    & \multicolumn{1}{c|}{Label} & \multicolumn{1}{c|}{0}     & \multicolumn{1}{c|}{1}     & \multicolumn{1}{c|}{2}     & \multicolumn{1}{c|}{3}     & \multicolumn{1}{c|}{4}     & \multicolumn{1}{c|}{0}     & \multicolumn{1}{c|}{1}     & \multicolumn{1}{c|}{2}     & \multicolumn{1}{c|}{3}     & \multicolumn{1}{c|}{4}    \\ \hline
\multicolumn{1}{|c|}{\multirow{4}{*}{ST}}   & \multicolumn{1}{c|}{\multirow{2}{*}{1}}   & Shortcut                   & \multicolumn{1}{c|}{-.851} & \multicolumn{1}{c|}{-.834} & \multicolumn{1}{c|}{-.582} & \multicolumn{1}{c|}{.416}  & 1.579 & \multicolumn{1}{c|}{-.587} & \multicolumn{1}{c|}{-.749} & \multicolumn{1}{c|}{-.736} & \multicolumn{1}{c|}{.421}  & 1.514 \\ \cline{3-13} 
\multicolumn{1}{|c|}{}                      & \multicolumn{1}{c|}{}                     & Others                     & \multicolumn{1}{c|}{-.007} & \multicolumn{1}{c|}{.003}  & \multicolumn{1}{c|}{.004}  & \multicolumn{1}{c|}{-.002} & -.007 & \multicolumn{1}{c|}{.002}  & \multicolumn{1}{c|}{.009}  & \multicolumn{1}{c|}{-.002} & \multicolumn{1}{c|}{.004}  & -.008 \\ \cline{2-13} 
\multicolumn{1}{|c|}{}                      & \multicolumn{1}{c|}{\multirow{2}{*}{0.8}} & Shortcut                   & \multicolumn{1}{c|}{-.231} & \multicolumn{1}{c|}{-.307} & \multicolumn{1}{c|}{-.178} & \multicolumn{1}{c|}{.156}  & .533  & \multicolumn{1}{c|}{-.238} & \multicolumn{1}{c|}{-.230} & \multicolumn{1}{c|}{-.265} & \multicolumn{1}{c|}{.209}  & .448  \\ \cline{3-13} 
\multicolumn{1}{|c|}{}                      & \multicolumn{1}{c|}{}                     & Others                     & \multicolumn{1}{c|}{.003}  & \multicolumn{1}{c|}{.008}  & \multicolumn{1}{c|}{.009}  & \multicolumn{1}{c|}{-.003} & -.012 & \multicolumn{1}{c|}{.008}  & \multicolumn{1}{c|}{.010}  & \multicolumn{1}{c|}{-.002} & \multicolumn{1}{c|}{.004}  & -.013 \\ \hline
\multicolumn{1}{|c|}{\multirow{4}{*}{Syn}}  & \multicolumn{1}{c|}{\multirow{2}{*}{1}}   & Shortcut                   & \multicolumn{1}{c|}{-.221} & \multicolumn{1}{c|}{-.241} & \multicolumn{1}{c|}{-.146} & \multicolumn{1}{c|}{.091}  & .332  & \multicolumn{1}{c|}{-.109} & \multicolumn{1}{c|}{-.177} & \multicolumn{1}{c|}{-.211} & \multicolumn{1}{c|}{.036}  & .400  \\ \cline{3-13} 
\multicolumn{1}{|c|}{}                      & \multicolumn{1}{c|}{}                     & Others                     & \multicolumn{1}{c|}{-.005} & \multicolumn{1}{c|}{.007}  & \multicolumn{1}{c|}{.005}  & \multicolumn{1}{c|}{-.001} & -.008 & \multicolumn{1}{c|}{.009}  & \multicolumn{1}{c|}{.011}  & \multicolumn{1}{c|}{-.004} & \multicolumn{1}{c|}{.002}  & -.009 \\ \cline{2-13} 
\multicolumn{1}{|c|}{}                      & \multicolumn{1}{c|}{\multirow{2}{*}{0.8}} & Shortcut                   & \multicolumn{1}{c|}{-.022} & \multicolumn{1}{c|}{-.056} & \multicolumn{1}{c|}{-.029} & \multicolumn{1}{c|}{.035}  & .080  & \multicolumn{1}{c|}{-.081} & \multicolumn{1}{c|}{-.117} & \multicolumn{1}{c|}{-.140} & \multicolumn{1}{c|}{.094}  & .210  \\ \cline{3-13} 
\multicolumn{1}{|c|}{}                      & \multicolumn{1}{c|}{}                     & Others                     & \multicolumn{1}{c|}{.000}  & \multicolumn{1}{c|}{.005}  & \multicolumn{1}{c|}{.004}  & \multicolumn{1}{c|}{-.001} & -.016 & \multicolumn{1}{c|}{.004}  & \multicolumn{1}{c|}{.006}  & \multicolumn{1}{c|}{-.009} & \multicolumn{1}{c|}{.007}  & -.003 \\ \hline
\multicolumn{1}{|c|}{\multirow{6}{*}{Catg}} & \multicolumn{1}{c|}{\multirow{3}{*}{1}}   & Country                    & \multicolumn{1}{c|}{-.106} & \multicolumn{1}{c|}{-.182} & \multicolumn{1}{c|}{.073}  & \multicolumn{1}{c|}{.015}  & .193  & \multicolumn{1}{c|}{-.100} & \multicolumn{1}{c|}{-.057} & \multicolumn{1}{c|}{-.010} & \multicolumn{1}{c|}{.080}  & .092  \\ \cline{3-13} 
\multicolumn{1}{|c|}{}                      & \multicolumn{1}{c|}{}                     & City                       & \multicolumn{1}{c|}{.095}  & \multicolumn{1}{c|}{.012}  & \multicolumn{1}{c|}{.047}  & \multicolumn{1}{c|}{-.021} & -.111 & \multicolumn{1}{c|}{.056}  & \multicolumn{1}{c|}{.052}  & \multicolumn{1}{c|}{.004}  & \multicolumn{1}{c|}{-.023} & -.080 \\ \cline{3-13} 
\multicolumn{1}{|c|}{}                      & \multicolumn{1}{c|}{}                     & Others                     & \multicolumn{1}{c|}{.000}  & \multicolumn{1}{c|}{.005}  & \multicolumn{1}{c|}{-.008} & \multicolumn{1}{c|}{.004}  & .000  & \multicolumn{1}{c|}{.005}  & \multicolumn{1}{c|}{.012}  & \multicolumn{1}{c|}{-.001} & \multicolumn{1}{c|}{.001}  & -.011 \\ \cline{2-13} 
\multicolumn{1}{|c|}{}                      & \multicolumn{1}{c|}{\multirow{3}{*}{0.8}} & Country                    & \multicolumn{1}{c|}{-.028} & \multicolumn{1}{c|}{-.050} & \multicolumn{1}{c|}{-.022} & \multicolumn{1}{c|}{-.069} & .160  & \multicolumn{1}{c|}{-.041} & \multicolumn{1}{c|}{-.033} & \multicolumn{1}{c|}{-.047} & \multicolumn{1}{c|}{.021}  & .106  \\ \cline{3-13} 
\multicolumn{1}{|c|}{}                      & \multicolumn{1}{c|}{}                     & City                       & \multicolumn{1}{c|}{.059}  & \multicolumn{1}{c|}{.016}  & \multicolumn{1}{c|}{-.006} & \multicolumn{1}{c|}{-.033} & -.047 & \multicolumn{1}{c|}{.029}  & \multicolumn{1}{c|}{.023}  & \multicolumn{1}{c|}{-.007} & \multicolumn{1}{c|}{-.020} & -.022 \\ \cline{3-13} 
\multicolumn{1}{|c|}{}                      & \multicolumn{1}{c|}{}                     & Others                     & \multicolumn{1}{c|}{.010}  & \multicolumn{1}{c|}{.012}  & \multicolumn{1}{c|}{.003}  & \multicolumn{1}{c|}{-.020} & -.009 & \multicolumn{1}{c|}{.004}  & \multicolumn{1}{c|}{.007}  & \multicolumn{1}{c|}{.000}  & \multicolumn{1}{c|}{.000}  & -.007 \\ \hline
\end{tabular}
\caption{SHAP values of BERT and AFR on Yelp. (ST = Single Term, Syn = Synonym, Catg = Category)}
\vspace{-15pt}
\label{tab:shap}
\end{table*}

\subsection{Model Analysis via Explainability}
From Section~\ref{sec: analysis-bert} to Section~\ref{sec: analysis-rm}, we have demonstrated that LMs, LLMs, and SOTA rubost models are all vulnerable to our proposed shortcuts. In this section, we further analyze the models' prediction behavior. Specifically, we use SHAP~\cite{lundberg2017unified} to analyze how the shortcut tokens affect the final prediction of the model.

Using the occurrence shortcuts as an example, we sampled 100 test instances from the Yelp test datasets containing shortcuts and calculated the average SHAP values for shortcut tokens and non-shortcut tokens for BERT and AFR. 

Table~\ref{tab:shap} shows the contributions of tokens to the model's prediction of each label. Single term shortcut tokens have relatively large positive SHAP values for label 4 and relatively large negative SHAP values for label 0, compared to non-shortcut tokens. This indicates that these tokens make the models more likely to predict label 4 and less likely to predict label 0, which aligns with the spurious correlations in training datasets. As the $\lambda$ decreases which means the shortcut strength decreases, those SHAP values also decrease, showing less impact of shortcuts on models' prediction. The same patterns also happen to synonym shortcuts. For the category shortcuts, we observe that the city category has a more positive impact on predicting label 0 and a negative impact on predicting label 4. In contrast, the country category has the opposite effect. This aligns with the spurious correlations in the training dataset, where the "country" is strongly associated with higher scores and the "city" with lower scores. 

\begin{figure}[t!]
    \centering
    \includegraphics[width=0.45\textwidth]{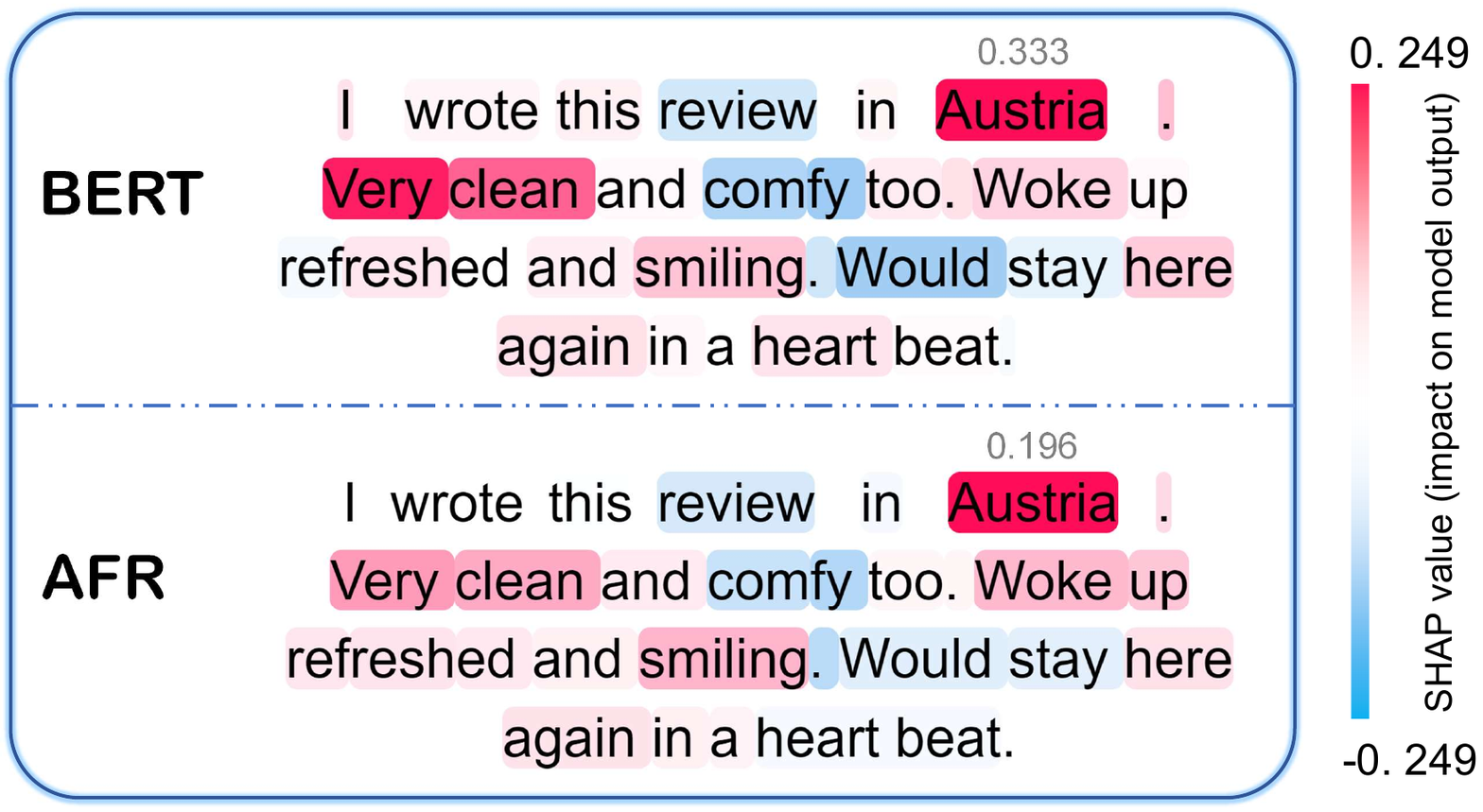}
    \vspace{-18pt}
    \caption{SHAP analysis for BERT and AFR.}
    \label{fig:shap_example}
    \vspace{-15pt}
\end{figure}

Besides, according to the SHAP values, AFR is more affected by synonym shortcuts, while BERT is more affected by category shortcuts. This is consistent with the observations in Figure~\ref{fig:emotions_f1_drop}. We present an example of SHAP analysis for BERT and AFR with a category shortcut in Figure~\ref{fig:shap_example}. Red indicates a positive contribution to predicting Label 4, while blue indicates a negative contribution.  In both models, the word "Austria" significantly influences the prediction, but its SHAP value is relatively smaller in AFR than in BERT.

\section{Related Work}
% \label{sec:related work}
With the tremendous progress of deep neural networks (DNNs) in various language and vision tasks, there is a growing interest within the community regarding the mechanisms of learning within these models and the features they have captured for prediction.  Some recent studies have uncovered a shortcut learning phenomenon of DNNs with the utilization of adversarial test sets~\cite{jia2017adversarial} and DNN explainability techniques~\cite{du2019techniques, wang2020score,deng2021unified}. These works reveal that DNNs tend to exploit spurious correlations, rather than focusing on higher-level, task-relevant features in the training data. This tendency can result in poor performance when generalizing to out-of-distribution samples. 

This phenomenon of shortcut learning is observed across various language and vision tasks, including NLI~\cite{niven2019probing}, question answering~\cite{mudrakarta2018did}, reading comprehension~\cite{si2019does}, and VQA~\cite{agrawal2018don, manjunatha2019explicit,si2022language}. For instance, \citet{du2021towards} found that BERT uses lexical bias as a shortcut in NLU tasks. They explained this shortcut learning behavior as a consequence of a long-tailed distribution. ~\citet{Tang2023LargeLM} also focuses on NLU tasks, aiming to determine whether LLMs resort to shortcut strategies in NLU tasks even without parameter updates. In their experiments, they designed several spurious correlations or shortcut patterns, embedding them into multiple input-label pairs as prompts. However, these studies only consider simple and limited features as shortcuts and mechanically insert spurious correlations without considering the semantics of the original text. In contrast, we formally and systematically define various types of shortcut triggers and integrate them into a dataset as seamlessly as possible, without altering the original meaning of the text.

\section{Conclusion}
In this paper, we aim to establish a benchmark for detecting shortcut learning behaviors in text classification tasks. We propose a series of definitions of text shortcuts, introduce a benchmark for LM robustness assessment against the defined shortcuts, and empirically demonstrate the susceptibility of BERT, Llama, and three SOTA robust models to occurrence, style, and concept-based shortcuts.

% \newpage
\section{Limitations}

The first limitation is that the tasks in our benchmark are limited to sentiment analysis and emotion prediction. We mainly focus on text classification without extending our benchmark to other capabilities of LMs. 

Another limitation lies in the use of LLM to rewrite text. Using the LLM to paraphrase text is time-consuming, making it impractical to rewrite entire original datasets. Limiting the output size can speed up the process, but this may truncate the text before fully conveying the original meaning, especially for longer texts. Conversely, setting a large output size to avoid truncation can lead to the LLM generating irrelevant text for shorter inputs, adding noise to the datasets. This creates a tradeoff. 

Third, we manually checked the quality of the datasets by sampling modified data and used GPT-4o~\cite{achiam2023gpt} to evaluate the quality across four criteria, assigning ratings for each. However, these methods are relatively basic. We do not yet have a more sophisticated or reliable approach for accurately assessing dataset quality. For instance, while we can intuitively detect different authors' writing styles, we lack effective methods to evaluate how closely the modified text matches the intended author's style. Additionally, it is possible that the rewrite process could cause a shift in the ground truth labels for some data, although this would not affect our conclusions. Our experiments assume that the powerful LLM will faithfully follow instructions, ensuring that a positive review does not become negative after rewriting. However, for medium reviews, there could be a label shift.

\section*{Ethics Statement}
All information presented in the modified datasets is fictional and any resemblance to actual locations, individuals, or events is purely coincidental. All contents in the datasets do NOT represent the authors' views.

% \section*{Acknowledgements}

\newpage

\bibliography{custom}

\appendix

\section{Appendices}
\label{sec:appendix}

\subsection{Datasets}
{\bf Yelp.} We randomly select 2000 samples for each rating from the original dataset, for both training and test sets. From this selected training set, we randomly chose 100 samples for each label to form the validation set. Finally, we get a basic Yelp dataset, in which the training set contains 9500 samples (1900 samples per label), the validation set contains 500 samples (100 samples per label) and the test set contains 10000 samples (2000 samples per label). Considering the Yelp dataset is too large to be modified when constructing style shortcuts, we randomly select 5000 samples (1000 per label) for both the training and the test sets, and split the training set into a validation set (500 samples in total, 100 per label) and a new training set (4500 samples).

{\bf Go Emotions.} The original Go Emotions dataset contains 28 emotions. For simplicity and to reflect the gradual intensification of emotions, we select the following four emotions: neutral, amusement, joy, and excitement. Finally, we get 1619 samples in the training set, 100 samples in the validation set, and 680 samples in the test set.

{\bf Beer.} The Beer dataset is a sentiment dataset containing different aspects. We use the Beer dataset from~\cite{bao2018deriving}. They collected three aspects: aroma, palate, and appearance. For each aspect, they gave token-level true rationales which are truly responsible for the rating of that aspect. For each aspect, we only keep the sentences that contain true rationales and then randomly choose 2000 samples whose corresponding ratings are in $\{0.4, 0.6, 0.8, 1.0\}$. The final base Beer dataset contains 2000 samples per aspect, for both training, test, and validation datasets, with ratings of $\{0.4, 0.6, 0.8, 1.0\}$.
\label{sec: datasets}

\subsection{Shortcuts Construction}

\subsubsection{Occurrence}

\textbf{Synonyms:}
\label{sec:appendix_syn}
The full set of the synonyms we used in our experiments contains
these phrases: "honestly", "to be honest", "frankly speaking", "to tell the truth", "to be frank", "in truth", "candidly", "speaking candidly", "plainly speaking", "to be direct", "to come clean", "to put it frankly", "if I’m being honest", "in plain terms", and "directly speaking".

\textbf{Category}
The list of country names is collected from \url{https://history.state.gov/countries/all}. The list of city names is mainly provided by ChatGPT-3.5~\cite{ouyang2022training}, and some cities are added or deleted manually.
The completed lists can refer to \url{https://github.com/yuqing-zhou/shortcut-learning-in-text-classification}.

\subsubsection{Style}
\label{sec:appendix_style}
\begin{figure}[h!]
    \centering
    \includegraphics[width=0.5\textwidth]{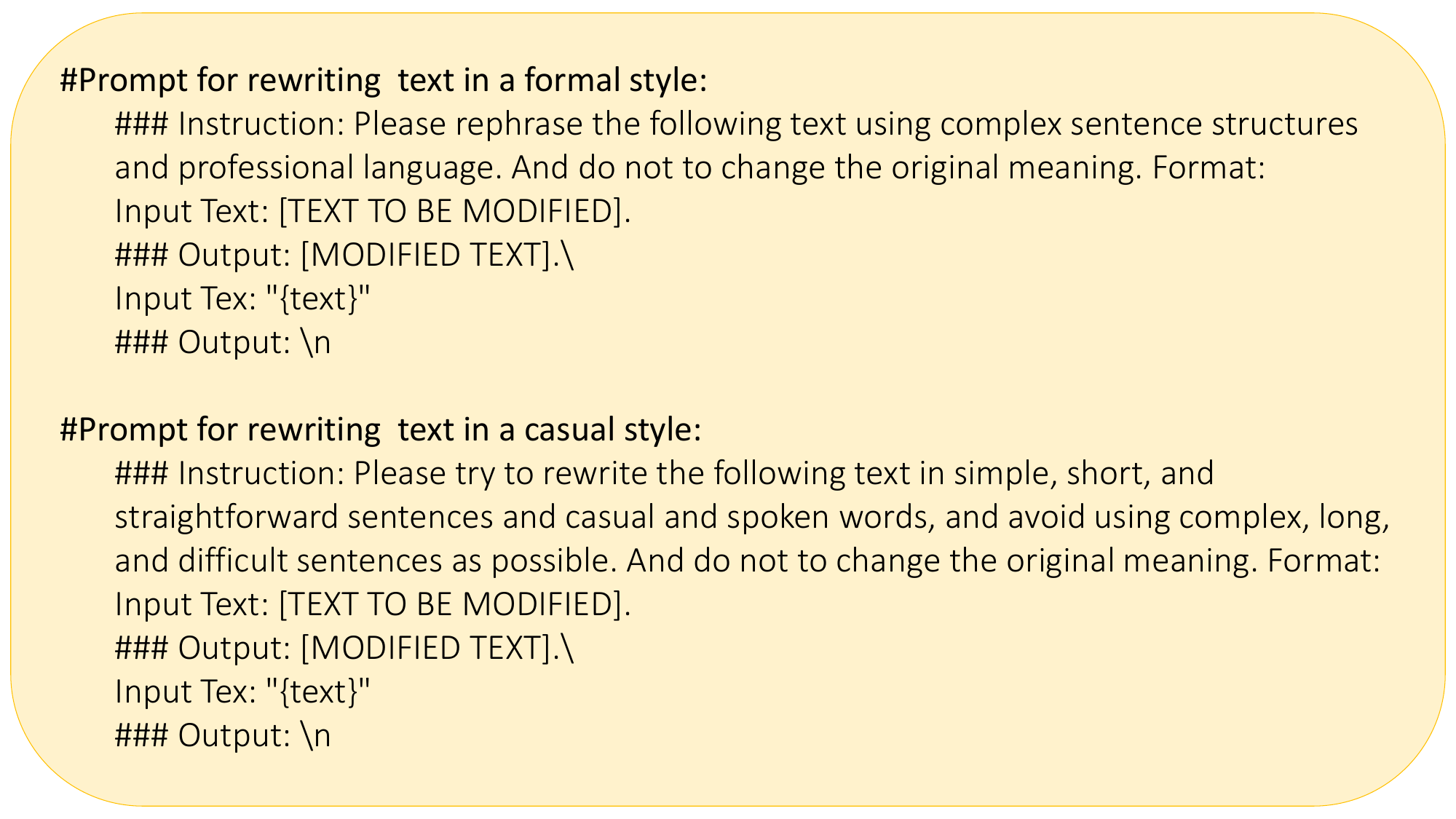}
    \caption{Prompts for generating texts with different register styles.}
    \label{fig:reg_prompt}
\end{figure}

\begin{figure}[ht!]
    \centering
    \includegraphics[width=0.5\textwidth]{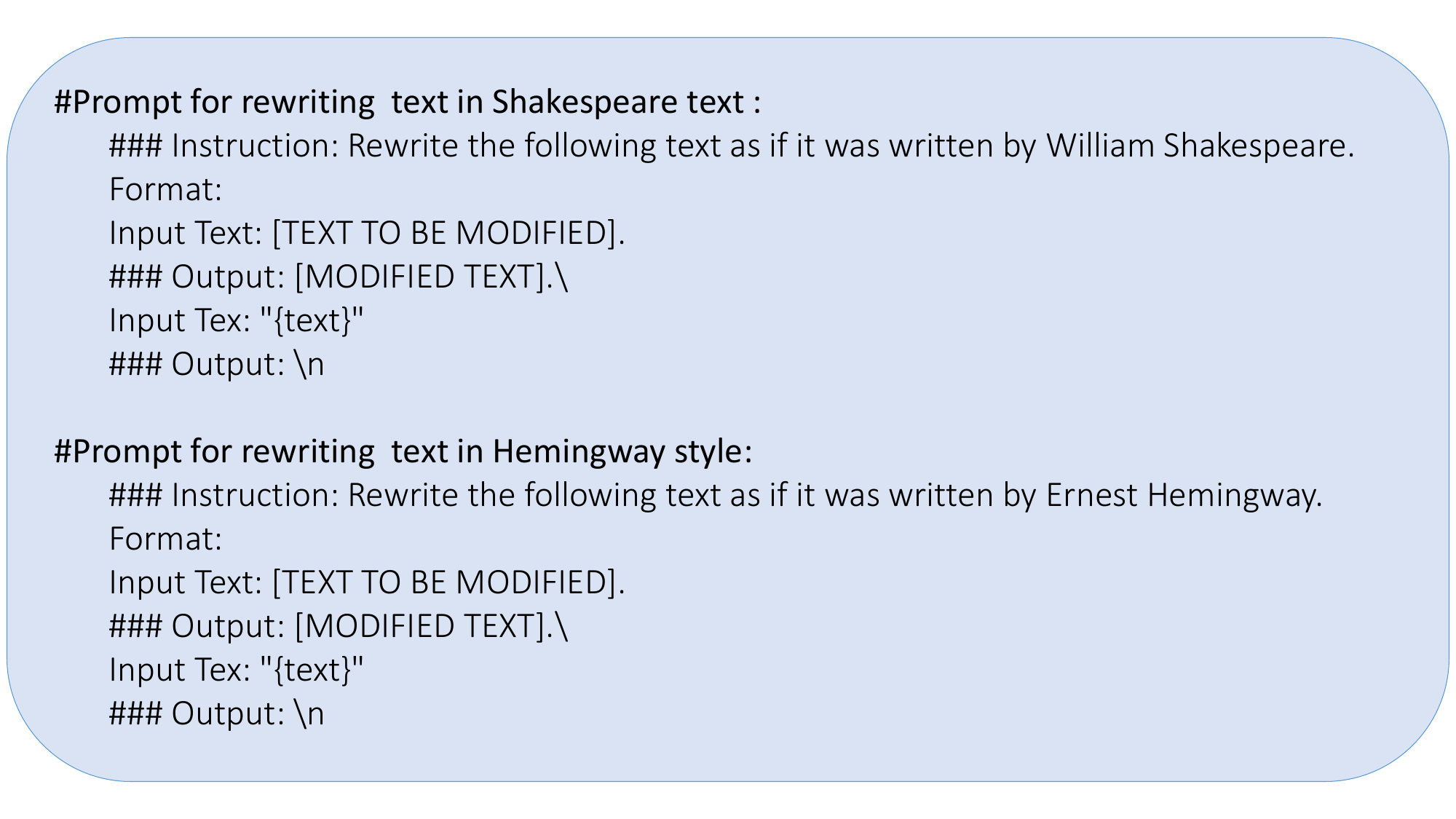}
    \caption{Prompts for generating texts with different author writing styles.}
    \label{fig:author_prompt}
\end{figure}

The prompts for transferring texts into different styles are shown in Figure~\ref{fig:reg_prompt} and ~\ref{fig:author_prompt}. 

\begin{figure*}[h!]
    \centering
    \includegraphics[width=1.0\textwidth]{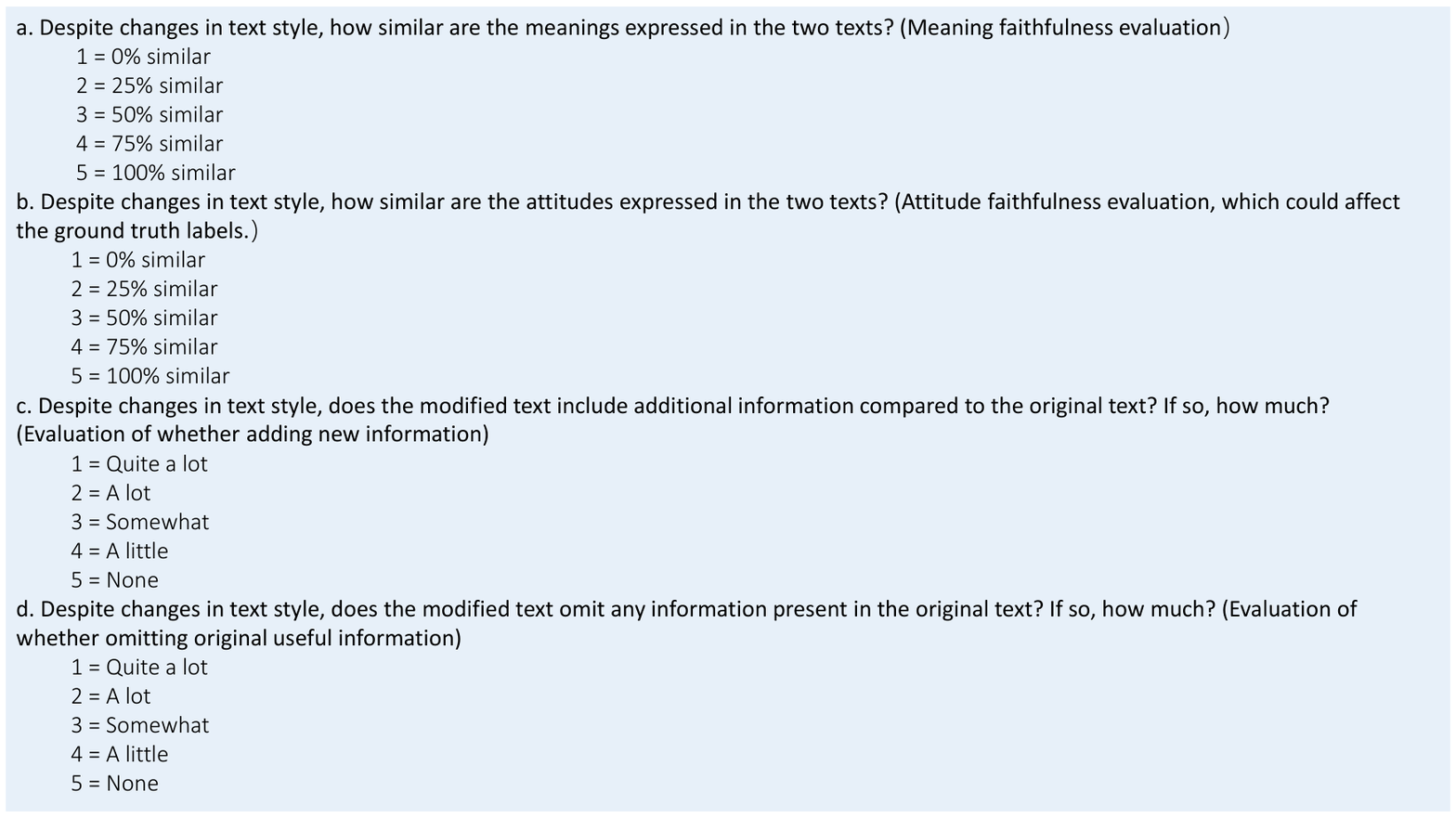}
    \vspace{-20pt}
    \caption{Criteria for evaluating dataset quality.}
    \label{fig:dataset_evaluate_criterion}
\end{figure*}

For evaluating the quality of the modified datasets, we used GPT-4 to assess them across four aspects, providing ratings for each criterion, as shown in Figure~\ref{fig:dataset_evaluate_criterion}. We then calculated the average score for all the samples. The results for the entire Yelp training dataset and the Go Emotions training dataset are shown in Table~\ref{table:quality_Yelp} and Table~\ref{table:quality_emo}, respectively. From the results, we can conclude that the modified datasets are faithful to the original dataset, as each metric achieves very high scores. The modified datasets demonstrate high quality.

\begin{table*}[ht!]\footnotesize
\begin{tabular}{|lcccc|}
\hline
\multicolumn{5}{|c|}{Yelp}                                                                                                                                                                            \\ \hline
\multicolumn{1}{|l|}{}            & \multicolumn{1}{l|}{Q1 (Meaning faithfulness)} & \multicolumn{1}{l|}{Q2 (Attitude faithfulness)} & \multicolumn{1}{l|}{Q3 (No added info)} & Q4 (No omitted info) \\ \hline
\multicolumn{1}{|l|}{Hemingway}   & \multicolumn{1}{c|}{5.00}                         & \multicolumn{1}{c|}{5.00}                          & \multicolumn{1}{c|}{5.00}                  & 4.00                   \\ \hline
\multicolumn{1}{|l|}{Shakespeare} & \multicolumn{1}{c|}{4.81}                      & \multicolumn{1}{c|}{4.12}                       & \multicolumn{1}{c|}{4.32}               & 4.69                 \\ \hline
\multicolumn{1}{|l|}{Formal}      & \multicolumn{1}{c|}{4.00}                         & \multicolumn{1}{c|}{4.10}                        & \multicolumn{1}{c|}{4.16}               & 4.84                 \\ \hline
\end{tabular}
\caption{Evaluation results of the modified Yelp datasets' quality.}
\label{table:quality_Yelp}
\end{table*}

\begin{table*}[ht!]\small
\begin{tabular}{|lcccc|}
\hline
\multicolumn{5}{|c|}{\textbf{Go   Emotions}}                                                                                                                                                                  \\ \hline
\multicolumn{1}{|l|}{}            & \multicolumn{1}{c|}{Q1 (Meaning   faithfulness)} & \multicolumn{1}{c|}{Q2 (Attitude   faithfulness)} & \multicolumn{1}{c|}{Q3 (No added   info)} & Q4 (No omitted   info) \\ \hline
\multicolumn{1}{|l|}{Hemingway}   & \multicolumn{1}{c|}{4.00}                           & \multicolumn{1}{c|}{4.04}                         & \multicolumn{1}{c|}{4.64}                 & 4.37                   \\ \hline
\multicolumn{1}{|l|}{Shakespeare} & \multicolumn{1}{c|}{4.00}                           & \multicolumn{1}{c|}{4.20}                          & \multicolumn{1}{c|}{4.01}                 & 4.99                   \\ \hline
\multicolumn{1}{|l|}{Formal}      & \multicolumn{1}{c|}{4.00}                           & \multicolumn{1}{c|}{4.05}                         & \multicolumn{1}{c|}{4.09}                 & 4.91                   \\ \hline
\end{tabular}
\caption{Evaluation results of the modified Go Emotions datasets' quality.}
\label{table:quality_emo}
\end{table*}

\begin{table}[]
\centering
\begin{tabular}{|l|c|} 
\hline
\textbf{Model} & \textbf{\#params} \\\hline
BERT           & 110M              \\\hline
A2R            & 2M                \\\hline
AFR            & 110M              \\\hline
CR             & 219M              \\\hline
Llama2-7b      & 7B                \\\hline
Llama3-8b      & 8B                \\\hline
Llama2-13b     & 13B              \\\hline
\end{tabular}
\caption{Model Size}
\label{tab:models_size}
\end{table}

\subsection{Experiment Setup and Results}
\label{sec:appendix_exp}
\subsubsection{BERT}
After the hyper-parameter search for the learning rate from $[2e-2, 2e-3, 2e-4, 2e-5]$, we choose the $2e-5$ as the learning rate for finetuning as it achieves the best performance. The weight decay is $0.01$ for Yelp and Beer and $0.1$ for Go Emotions. The batch size is $16$. The number of training epochs is $15$ for the Yelp and Go Emotions datasets and $20$ for the Beer dataset. The overall performances of BERT are shown in Table~\ref{tab:bert-acc} and Table~\ref{tab:bert-f1}.
\label{sec: bert-setting}

\subsubsection{LLMs}
We use QLoRA~\cite{dettmers2023qlora} to finetune LLMs. We searched the best learning rate from $[2e-5, 2e-4, 2e-3, 2e-2, 2e-1]$, lora rank from $[64, 128, 256]$, and lora alpha from  $[8, 16, 32]$. We set the learning rate as $5e-4$ for Yelp, $2e-3$ for Go Emotions and Beer. The number of epochs is 4 for Go Emotions, 8 for Beer, and 10 for Yelp. The parameters of QLoRA can refer to "$code/utils.py: Hyperparameter$" in \url{https://github.com/yuqing-zhou/shortcut-learning-in-text-classification}.

The settings for the LLM inference phase are as follows:
\begin{quote}
\begin{verbatim}
 return_full_text=True, 
 task='text-generation',
 temperature= 0.0000, 
 max_new_tokens=5, 
 repetition_penalty=1.1.
\end{verbatim}
\end{quote}
\begin{figure}[h!]
    \centering
    \includegraphics[width=0.5\textwidth]{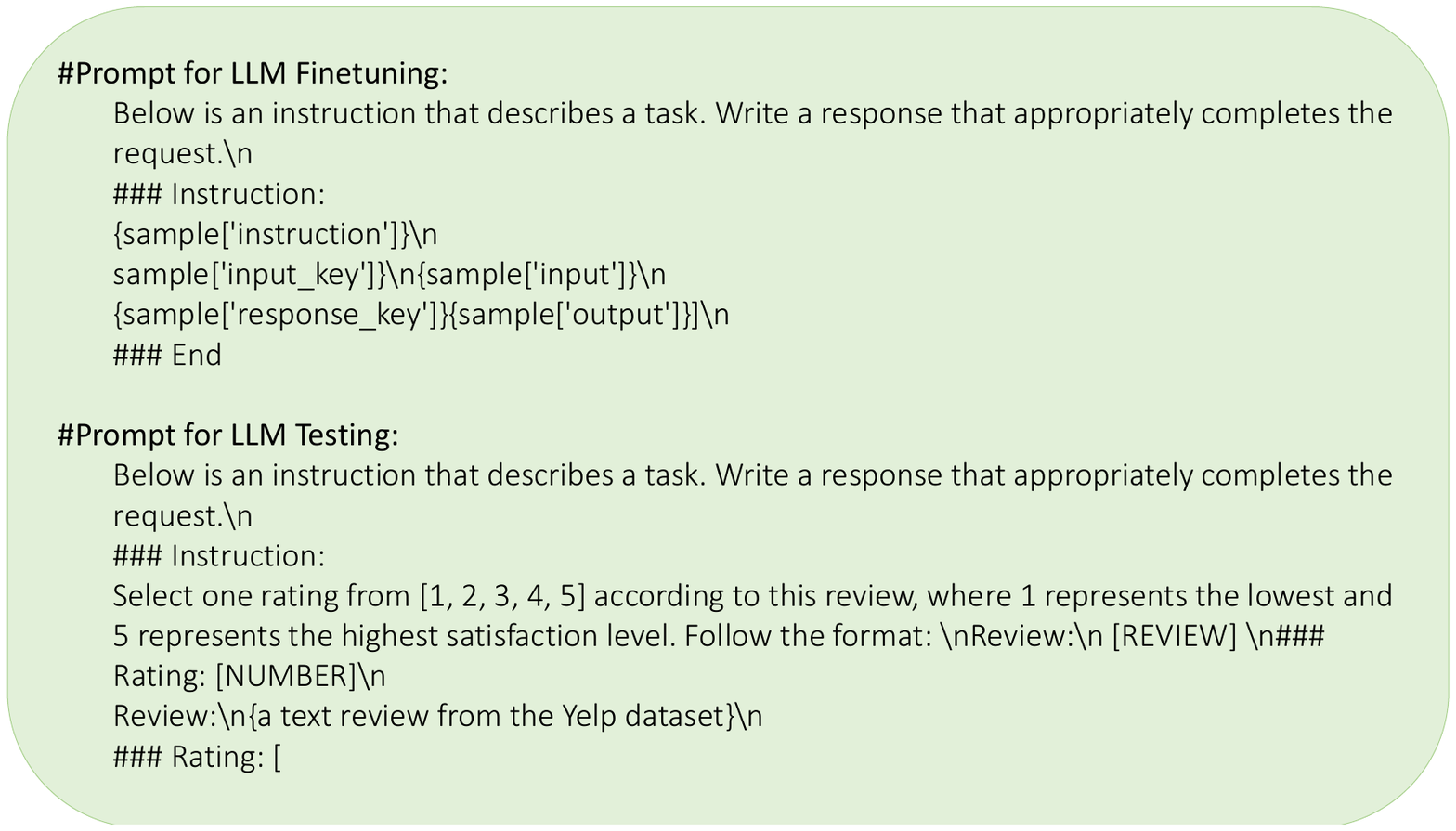}
    \caption{Prompts for LLM finetuning and evaluation.}
    \label{fig:llm_finetune_test_prompt}
\end{figure}
The prompts for LLM fine-tuning and evaluation are shown in Figure~\ref{fig:llm_finetune_test_prompt}. The fields "instruction", "input\_key", "input", "response\_key", and "output" in the prompts vary for each dataset, as defined in our code file $utils.py/DATASETS\_INFO$. The output of the LLM evaluation phase is a string. We extract the first output character after the sequence "$\#\#\# Rating: [$" and convert it to an integer as the rating, then compare it with ground truth for evaluation.

The overall performances of Llama2-13b are shown in Table~\ref{tab:llama2-13b-acc} and Table~\ref{tab:llama2-13b-f1}.
\label{sec: llama-setting}

\subsubsection{A2R}
The setting of hyperparameters can refer to "$code/beer\_data\_utils\_neurips21.py$" in \url{https://github.com/yuqing-zhou/shortcut-learning-in-text-classification}.

The overall performances of A2R are shown in Table~\ref{tab:a2r-acc} and Table~\ref{tab:a2r-f1}.

\subsubsection{CR}
After briefly trying out a few hyper-parameters, for Yelp, Go Emotions, and Beer, we set the learning rates as $2e-5$, $5e-5$, and $2e-5$, respectively. The number of training epochs for Yelp, Go Emotions, and Beer is set as $8$, $15$, and $20$ respectively. The batch size is $16$. The overall performances of CR are shown in Table~\ref{tab:cr-acc} and Table~\ref{tab:cr-f1}.
\label{sec: cr-setting}

\subsubsection{AFR}
The learning rate is $1e-5$ except for Go Emotions with single term and synonym shortcuts, which is $1e-4$. Other hyperparameters can refer to can refer to "$code/AFRmodel.py$" in \url{https://github.com/yuqing-zhou/shortcut-learning-in-text-classification}. The overall performances of CR are shown in Table~\ref{tab:afr-acc} and Table~\ref{tab:afr-f1}.

\begin{figure*}[htbp]
    \centering
    \begin{subfigure}[b]{0.4\textwidth}
        \centering
        \includegraphics[width=\textwidth]{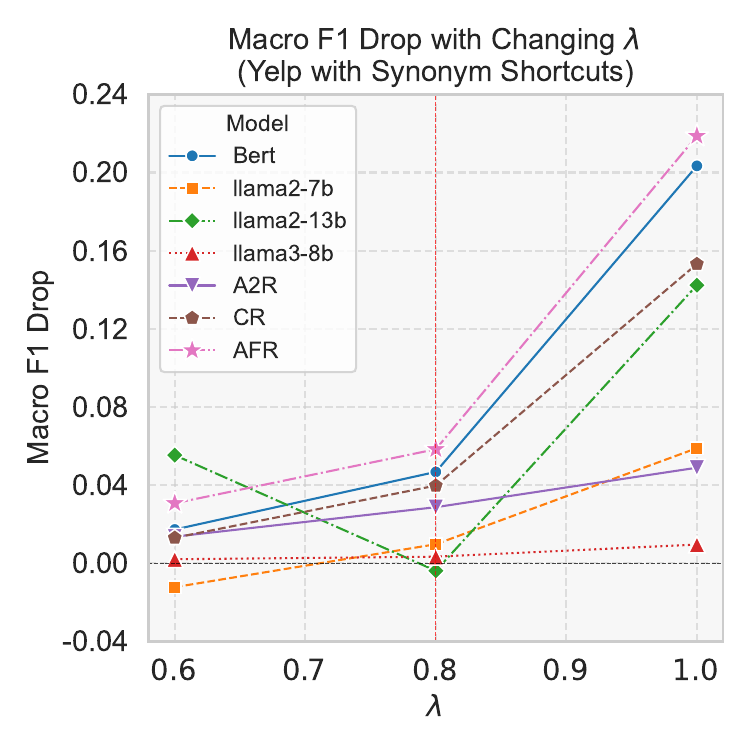}
        \caption{Synonym Shortcuts}
        \label{fig:yelp_syn_f1_drop}
    \end{subfigure}
    \hfill
    \begin{subfigure}[b]{0.4\textwidth}
        \centering
        \includegraphics[width=\textwidth]{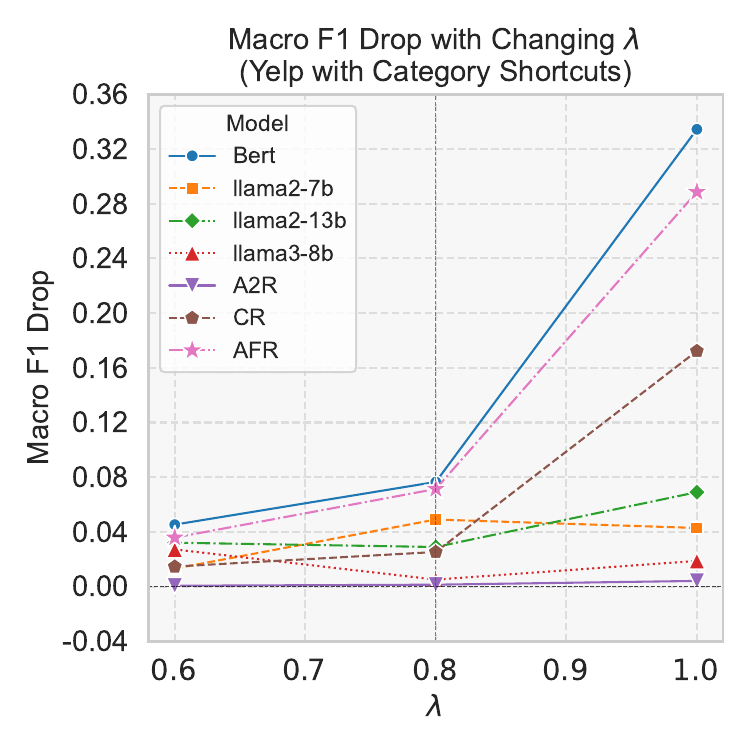}
        \caption{Category Shortcuts}
        \label{fig:yelp_catg_f1_drop}
    \end{subfigure}
    \hfill
    \begin{subfigure}[b]{0.4\textwidth}
        \centering
        \includegraphics[width=\textwidth]{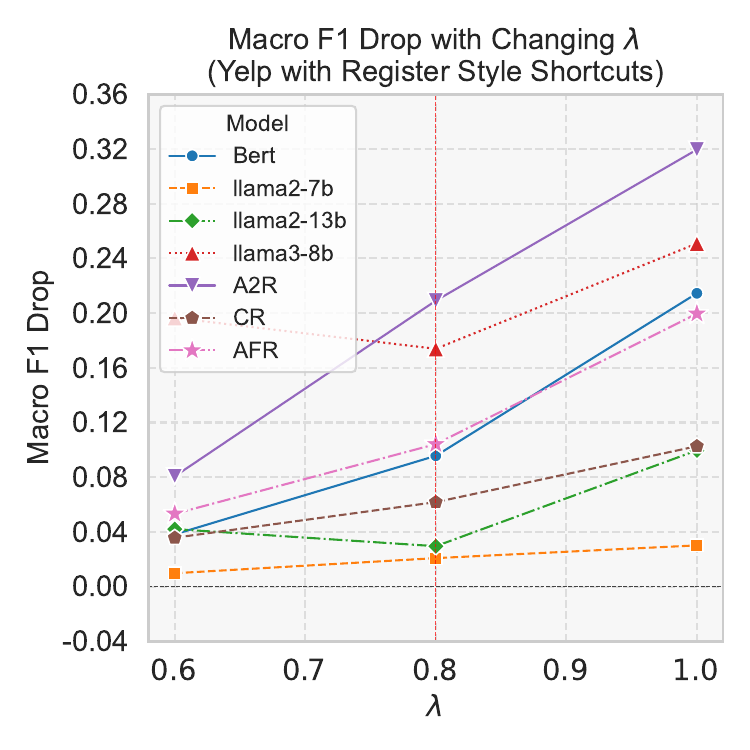}
        \caption{Register Style Shortcuts}
        \label{fig:yelp_reg_f1_drop}
    \end{subfigure}
    \hfill
    \begin{subfigure}[b]{0.4\textwidth}
        \centering
        \includegraphics[width=\textwidth]{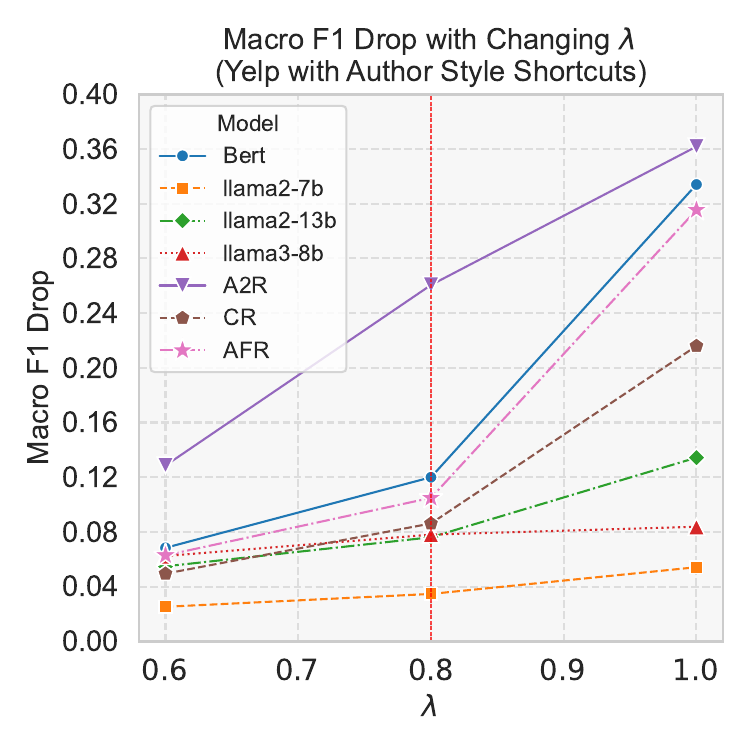}
        \caption{Author Style Shortcuts}
        \label{fig:yelp_auth_f1_drop}
    \end{subfigure}
    \caption{Macro F1 Drop with $\lambda$ (Yelp)}
    \vspace{-15pt}
    \label{fig:yelp_f1_drop}
\end{figure*}

\begin{figure*}[htbp]
    \centering
    \begin{subfigure}[b]{0.4\textwidth}
        \centering
        \includegraphics[width=\textwidth]{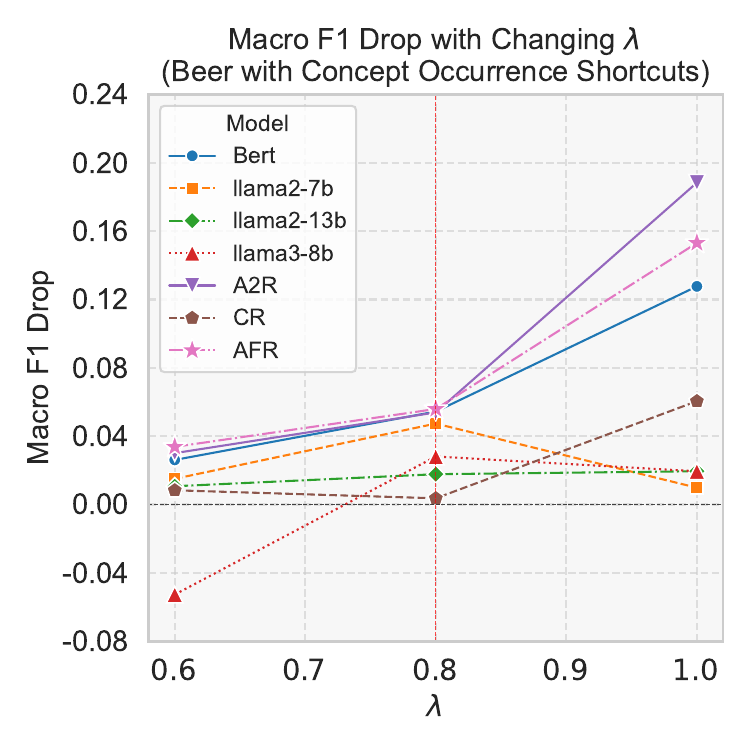}
        \caption{Concept Occurrence Shortcuts}
        \label{fig:beer_occur_f1_drop}
    \end{subfigure}
    \hfill
    \begin{subfigure}[b]{0.4\textwidth}
        \centering
        \includegraphics[width=\textwidth]{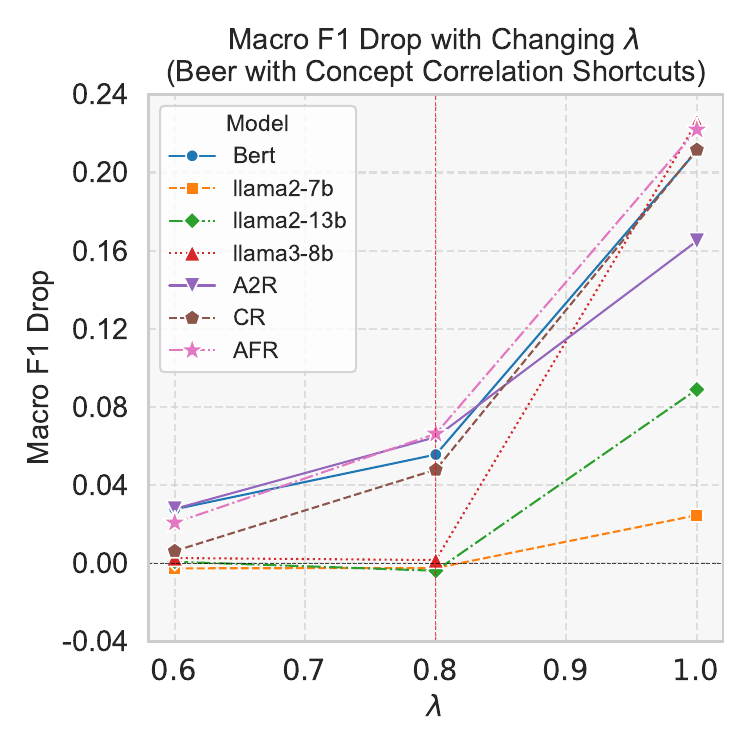}
        \caption{Concept Correlation Shortcuts}
        \label{fig:beer_corr_f1_drop}
    \end{subfigure}
    \caption{Macro F1 Drop with $\lambda$ (Beer)}
    \vspace{-15pt}
    \label{fig:beer_f1_drop}
\end{figure*}

\begin{table*}[t!] \small
\centering
\begin{tabular}{|ccl|ccc|ccc|ccc|}
\hline
\multicolumn{3}{|c|}{\textbf{Models}}                                                                   & \multicolumn{3}{c|}{\textbf{A2R}}                                                                        & \multicolumn{3}{c|}{\textbf{CR}}                                                                         & \multicolumn{3}{c|}{\textbf{AFR}}                                                                        \\ \hline
\multicolumn{1}{|c|}{\textbf{Datasets}}         & \multicolumn{2}{c|}{\textbf{Shortcut Types}}          & \multicolumn{1}{c|}{\textbf{Test}} & \multicolumn{1}{c|}{\textbf{Anti}} & \multicolumn{1}{c|}{\textbf{$\Delta$}} & \multicolumn{1}{c|}{\textbf{Test}} & \multicolumn{1}{c|}{\textbf{Anti}} & \multicolumn{1}{c|}{\textbf{$\Delta$}} & \multicolumn{1}{c|}{\textbf{Test}} & \multicolumn{1}{c|}{\textbf{Anti}} & \multicolumn{1}{c|}{\textbf{$\Delta$}} \\ \hline
\multicolumn{1}{|c|}{\multirow{5}{*}{Yelp}}     & \multicolumn{1}{c|}{\multirow{3}{*}{Occur}}   & ST    & \multicolumn{1}{c|}{.533}          & \multicolumn{1}{c|}{.436}          & \textbf{.097}                  & \multicolumn{1}{c|}{.555}          & \multicolumn{1}{c|}{.448}          & \textbf{.107}                  & \multicolumn{1}{c|}{.641}          & \multicolumn{1}{c|}{.369}          & \textbf{.272}                  \\ \cline{3-12} 
\multicolumn{1}{|c|}{}                          & \multicolumn{1}{c|}{}                         & Syn   & \multicolumn{1}{c|}{.507}          & \multicolumn{1}{c|}{.457}          & \textbf{.050}                  & \multicolumn{1}{c|}{.545}          & \multicolumn{1}{c|}{.462}          & \textbf{.083}                  & \multicolumn{1}{c|}{.642}          & \multicolumn{1}{c|}{.439}          & \textbf{.204}                  \\
\hline
\end{tabular}

\caption{Accuracy of three robust models, A2R, CR, and AFR. (We use the following abbreviations: Anti=Anti-test, Occur=Occurrence, ST=Single Term, Syn=Synonym, Catg=Category, Reg=Register, Auth=Author, Corr=Correlation.) All robust models show a statistically significant decrease in accuracy with $p<0.05$.}
\label{tab:rm-results-acc}
\end{table*}

% \usepackage{multirow}
\begin{table*}[]\small
\centering
\begin{tabular}{|cc|lccccccccc|}
\hline
\multicolumn{1}{|l|}{\textbf{}}                             & \multicolumn{1}{l|}{\textbf{}}     & \multicolumn{10}{c|}{\textbf{BERT   - Accuracy}}                                                                                                                                                                                                                                                                                                  \\ \hline
\multicolumn{2}{|c|}{\textbf{Datasets}}                                                          & \multicolumn{1}{l|}{\textbf{}}           & \multicolumn{3}{c|}{\textbf{Yelp}}                                                                      & \multicolumn{3}{c|}{\textbf{Go Emotions}}                                                               & \multicolumn{3}{c|}{\textbf{Beer}}                                                 \\ \hline
\multicolumn{2}{|c|}{\textbf{Shortcut}}                                                          & \multicolumn{1}{l|}{\textbf{$\lambda$}}     & \multicolumn{1}{c|}{\textbf{1}} & \multicolumn{1}{c|}{\textbf{0.8}} & \multicolumn{1}{c|}{\textbf{0.6}} & \multicolumn{1}{c|}{\textbf{1}} & \multicolumn{1}{c|}{\textbf{0.8}} & \multicolumn{1}{c|}{\textbf{0.6}} & \multicolumn{1}{c|}{\textbf{1}} & \multicolumn{1}{c|}{\textbf{0.8}} & \textbf{0.6} \\ \hline
\multicolumn{1}{|c|}{\multirow{15}{*}{\textbf{Occurrence}}} & \multirow{5}{*}{\textbf{ST}}       & \multicolumn{1}{l|}{\textbf{Test}}       & \multicolumn{1}{c|}{.634}       & \multicolumn{1}{c|}{.628}         & \multicolumn{1}{c|}{.614}         & \multicolumn{1}{c|}{.914}       & \multicolumn{1}{c|}{.802}         & \multicolumn{1}{c|}{.865}         & \multicolumn{1}{c|}{-}          & \multicolumn{1}{c|}{-}            & -            \\
\hline
\end{tabular}
\caption{Overall performances of BERT (Accuracy) ("Ori-Test" denotes the results on the original, unmodified test datasets. VAR($\Delta$) represents the variance of $\Delta$ across multiple experiments.)}
\label{tab:bert-acc}
\end{table*}

\begin{table*}[]\small
\begin{tabular}{|cc|lllllllccc|}
\hline
\multicolumn{1}{|l|}{\textbf{}}                             & \multicolumn{1}{l|}{\textbf{}}     & \multicolumn{10}{c|}{\textbf{BERT   - Macro F1}}                                                                                                                                                                                                                                                                                                                \\ \hline
\multicolumn{2}{|c|}{\textbf{Datasets}}                                                          & \multicolumn{1}{l|}{\textbf{}}           & \multicolumn{3}{c|}{\textbf{Yelp}}                                                                      & \multicolumn{3}{c|}{\textbf{Go Emotions}}                                                               & \multicolumn{3}{c|}{\textbf{Beer}}                                                               \\ \hline
\multicolumn{2}{|c|}{\textbf{Shortcut}}                                                          & \multicolumn{1}{l|}{\textbf{$\lambda$}}     & \multicolumn{1}{c|}{\textbf{1}} & \multicolumn{1}{c|}{\textbf{0.8}} & \multicolumn{1}{c|}{\textbf{0.6}} & \multicolumn{1}{c|}{\textbf{1}} & \multicolumn{1}{c|}{\textbf{0.8}} & \multicolumn{1}{c|}{\textbf{0.6}} & \multicolumn{1}{c|}{\textbf{1}} & \multicolumn{1}{c|}{\textbf{0.8}} & \textbf{0.6}               \\ \hline
\multicolumn{1}{|c|}{\multirow{15}{*}{\textbf{Occurrence}}} & \multirow{5}{*}{\textbf{ST}}       & \multicolumn{1}{l|}{\textbf{Test}}       & \multicolumn{1}{l|}{0.625}      & \multicolumn{1}{l|}{0.628}        & \multicolumn{1}{l|}{0.609}        & \multicolumn{1}{l|}{0.834}      & \multicolumn{1}{l|}{0.695}        & \multicolumn{1}{l|}{0.725}        & \multicolumn{1}{c|}{-}          & \multicolumn{1}{c|}{-}            & -                          \\
\hline
\end{tabular}
\caption{Overall performances of BERT (Macro F1)}
\label{tab:bert-f1}
\end{table*}

% \usepackage{multirow}
\begin{table*}[]\small
\begin{tabular}{|cc|lccccccccc|}
\hline
\multicolumn{1}{|l|}{\textbf{}}                             & \multicolumn{1}{l|}{\textbf{}}     & \multicolumn{10}{c|}{\textbf{Llama2-7b - Accuracy}}                                                                                                                                                                                                                                                                                               \\ \hline
\multicolumn{2}{|c|}{\textbf{Datasets}}                                                          & \multicolumn{1}{l|}{\textbf{}}           & \multicolumn{3}{c|}{\textbf{Yelp}}                                                                      & \multicolumn{3}{c|}{\textbf{Go Emotions}}                                                               & \multicolumn{3}{c|}{\textbf{Beer}}                                                 \\ \hline
\multicolumn{2}{|c|}{\textbf{Shortcut}}                                                          & \multicolumn{1}{l|}{\textbf{$\lambda$}}     & \multicolumn{1}{c|}{\textbf{1}} & \multicolumn{1}{c|}{\textbf{0.8}} & \multicolumn{1}{c|}{\textbf{0.6}} & \multicolumn{1}{c|}{\textbf{1}} & \multicolumn{1}{c|}{\textbf{0.8}} & \multicolumn{1}{c|}{\textbf{0.6}} & \multicolumn{1}{c|}{\textbf{1}} & \multicolumn{1}{c|}{\textbf{0.8}} & \textbf{0.6} \\ \hline
\multicolumn{1}{|c|}{\multirow{15}{*}{\textbf{Occurrence}}} & \multirow{5}{*}{\textbf{ST}}       & \multicolumn{1}{l|}{\textbf{Test}}       & \multicolumn{1}{c|}{0.719}      & \multicolumn{1}{c|}{0.707}        & \multicolumn{1}{c|}{0.702}        & \multicolumn{1}{c|}{0.868}      & \multicolumn{1}{c|}{0.851}        & \multicolumn{1}{c|}{0.829}        & \multicolumn{1}{c|}{-}          & \multicolumn{1}{c|}{-}            & -            \\
\hline
\end{tabular}
\caption{Overall performances of Llama2-7b (Accuracy)}
\label{tab:llama2-7b-acc}
\end{table*}

\begin{table*}[]\small
\centering
\begin{tabular}{|cc|lccccccccc|}
\hline
\multicolumn{1}{|l|}{\textbf{}}                             & \multicolumn{1}{l|}{\textbf{}}     & \multicolumn{10}{c|}{\textbf{Llama2-7b - Macro F1}}                                                                                                                                                                                                                                                                                               \\ \hline
\multicolumn{2}{|c|}{\textbf{Datasets}}                                                          & \multicolumn{1}{l|}{\textbf{}}           & \multicolumn{3}{c|}{\textbf{Yelp}}                                                                      & \multicolumn{3}{c|}{\textbf{Go Emotions}}                                                               & \multicolumn{3}{c|}{\textbf{Beer}}                                                 \\ \hline
\multicolumn{2}{|c|}{\textbf{Shortcut}}                                                          & \multicolumn{1}{l|}{\textbf{$\lambda$}}     & \multicolumn{1}{c|}{\textbf{1}} & \multicolumn{1}{c|}{\textbf{0.8}} & \multicolumn{1}{c|}{\textbf{0.6}} & \multicolumn{1}{c|}{\textbf{1}} & \multicolumn{1}{c|}{\textbf{0.8}} & \multicolumn{1}{c|}{\textbf{0.6}} & \multicolumn{1}{c|}{\textbf{1}} & \multicolumn{1}{c|}{\textbf{0.8}} & \textbf{0.6} \\ \hline
\multicolumn{1}{|c|}{\multirow{15}{*}{\textbf{Occurrence}}} & \multirow{5}{*}{\textbf{ST}}       & \multicolumn{1}{l|}{\textbf{Test}}       & \multicolumn{1}{c|}{0.514}      & \multicolumn{1}{c|}{0.551}        & \multicolumn{1}{c|}{0.531}        & \multicolumn{1}{c|}{0.761}      & \multicolumn{1}{c|}{0.723}        & \multicolumn{1}{c|}{0.715}        & \multicolumn{1}{c|}{-}          & \multicolumn{1}{c|}{-}            & -            \\
\hline
\end{tabular}
\caption{Overall performances of Llama2-7b (Macro F1)}
\label{tab:llama2-7b-f1}
\end{table*}

\begin{table*}[]\small
\centering
\begin{tabular}{|cc|lccccccccc|}
\hline
\multicolumn{1}{|l|}{\textbf{}}                             & \multicolumn{1}{l|}{\textbf{}}     & \multicolumn{10}{c|}{\textbf{Llama2-13b - Accuracy}}                                                                                                                                                                                                                                                                                              \\ \hline
\multicolumn{2}{|c|}{\textbf{Datasets}}                                                          & \multicolumn{1}{l|}{\textbf{}}           & \multicolumn{3}{c|}{\textbf{Yelp}}                                                                      & \multicolumn{3}{c|}{\textbf{Go Emotions}}                                                               & \multicolumn{3}{c|}{\textbf{Beer}}                                                 \\ \hline
\multicolumn{2}{|c|}{\textbf{Shortcut}}                                                          & \multicolumn{1}{l|}{\textbf{$\lambda$}}     & \multicolumn{1}{c|}{\textbf{1}} & \multicolumn{1}{c|}{\textbf{0.8}} & \multicolumn{1}{c|}{\textbf{0.6}} & \multicolumn{1}{c|}{\textbf{1}} & \multicolumn{1}{c|}{\textbf{0.8}} & \multicolumn{1}{c|}{\textbf{0.6}} & \multicolumn{1}{c|}{\textbf{1}} & \multicolumn{1}{c|}{\textbf{0.8}} & \textbf{0.6} \\ \hline
\multicolumn{1}{|c|}{\multirow{15}{*}{\textbf{Occurrence}}} & \multirow{5}{*}{\textbf{ST}}       & \multicolumn{1}{l|}{\textbf{Test}}       & \multicolumn{1}{c|}{0.727}      & \multicolumn{1}{c|}{0.712}        & \multicolumn{1}{c|}{0.698}        & \multicolumn{1}{c|}{0.761}      & \multicolumn{1}{c|}{0.764}        & \multicolumn{1}{c|}{0.806}        & \multicolumn{1}{c|}{-}          & \multicolumn{1}{c|}{-}            & -            \\
\hline
\end{tabular}
\caption{Overall performances of Llama2-13b (Accuracy)}
\label{tab:llama2-13b-acc}
\end{table*}

\begin{table*}[]\small
\centering
\begin{tabular}{|cc|lccccccccc|}
\hline
\multicolumn{1}{|l|}{\textbf{}}                             & \multicolumn{1}{l|}{\textbf{}}     & \multicolumn{10}{c|}{\textbf{Llama2-13b - Macro F1}}                                                                                                                                                                                                                                                                                              \\ \hline
\multicolumn{2}{|c|}{\textbf{Datasets}}                                                          & \multicolumn{1}{l|}{\textbf{}}           & \multicolumn{3}{c|}{\textbf{Yelp}}                                                                      & \multicolumn{3}{c|}{\textbf{Go Emotions}}                                                               & \multicolumn{3}{c|}{\textbf{Beer}}                                                 \\ \hline
\multicolumn{2}{|c|}{\textbf{Shortcut}}                                                          & \multicolumn{1}{l|}{\textbf{$\lambda$}}     & \multicolumn{1}{c|}{\textbf{1}} & \multicolumn{1}{c|}{\textbf{0.8}} & \multicolumn{1}{c|}{\textbf{0.6}} & \multicolumn{1}{c|}{\textbf{1}} & \multicolumn{1}{c|}{\textbf{0.8}} & \multicolumn{1}{c|}{\textbf{0.6}} & \multicolumn{1}{c|}{\textbf{1}} & \multicolumn{1}{c|}{\textbf{0.8}} & \textbf{0.6} \\ \hline
\multicolumn{1}{|c|}{\multirow{15}{*}{\textbf{Occurrence}}} & \multirow{5}{*}{\textbf{ST}}       & \multicolumn{1}{l|}{\textbf{Test}}       & \multicolumn{1}{c|}{0.723}      & \multicolumn{1}{c|}{0.507}        & \multicolumn{1}{c|}{0.520}        & \multicolumn{1}{c|}{0.675}      & \multicolumn{1}{c|}{0.628}        & \multicolumn{1}{c|}{0.723}        & \multicolumn{1}{c|}{-}          & \multicolumn{1}{c|}{-}            & -            \\
\hline
\end{tabular}
\caption{Overall performances of Llama2-13b (Macro F1)}
\label{tab:llama2-13b-f1}
\end{table*}

\begin{table*}[]\small
\centering
\begin{tabular}{|cc|lccccccccc|}
\hline
\multicolumn{1}{|l|}{\textbf{}}                             & \multicolumn{1}{l|}{\textbf{}}     & \multicolumn{10}{c|}{\textbf{Llama3-8b - Accuracy}}                                                                                                                                                                                                                                                                                               \\ \hline
\multicolumn{2}{|c|}{\textbf{Datasets}}                                                          & \multicolumn{1}{l|}{\textbf{}}           & \multicolumn{3}{c|}{\textbf{Yelp}}                                                                      & \multicolumn{3}{c|}{\textbf{Go Emotions}}                                                               & \multicolumn{3}{c|}{\textbf{Beer}}                                                 \\ \hline
\multicolumn{2}{|c|}{\textbf{Shortcut}}                                                          & \multicolumn{1}{l|}{\textbf{$\lambda$}}     & \multicolumn{1}{c|}{\textbf{1}} & \multicolumn{1}{c|}{\textbf{0.8}} & \multicolumn{1}{c|}{\textbf{0.6}} & \multicolumn{1}{c|}{\textbf{1}} & \multicolumn{1}{c|}{\textbf{0.8}} & \multicolumn{1}{c|}{\textbf{0.6}} & \multicolumn{1}{c|}{\textbf{1}} & \multicolumn{1}{c|}{\textbf{0.8}} & \textbf{0.6} \\ \hline
\multicolumn{1}{|c|}{\multirow{15}{*}{\textbf{Occurrence}}} & \multirow{5}{*}{\textbf{ST}}       & \multicolumn{1}{l|}{\textbf{Test}}       & \multicolumn{1}{c|}{0.726}      & \multicolumn{1}{c|}{0.719}        & \multicolumn{1}{c|}{0.703}        & \multicolumn{1}{c|}{0.839}      & \multicolumn{1}{c|}{0.835}        & \multicolumn{1}{c|}{0.820}        & \multicolumn{1}{c|}{-}          & \multicolumn{1}{c|}{-}            & -            \\
\hline
\end{tabular}
\caption{Overall performances of Llama3-8b (Accuracy)}
\label{tab:llama3-8b-acc}
\end{table*}

\begin{table*}[]\small
\begin{tabular}{|cc|lccccccccc|}
\hline
\multicolumn{1}{|l|}{\textbf{}}                             & \multicolumn{1}{l|}{\textbf{}}     & \multicolumn{10}{c|}{\textbf{Llama3-8b - Macro F1}}                                                                                                                                                                                                                                                                                               \\ \hline
\multicolumn{2}{|c|}{\textbf{Datasets}}                                                          & \multicolumn{1}{l|}{\textbf{}}           & \multicolumn{3}{c|}{\textbf{Yelp}}                                                                      & \multicolumn{3}{c|}{\textbf{Go Emotions}}                                                               & \multicolumn{3}{c|}{\textbf{Beer}}                                                 \\ \hline
\multicolumn{2}{|c|}{\textbf{Shortcut}}                                                          & \multicolumn{1}{l|}{\textbf{$\lambda$}}     & \multicolumn{1}{c|}{\textbf{1}} & \multicolumn{1}{c|}{\textbf{0.8}} & \multicolumn{1}{c|}{\textbf{0.6}} & \multicolumn{1}{c|}{\textbf{1}} & \multicolumn{1}{c|}{\textbf{0.8}} & \multicolumn{1}{c|}{\textbf{0.6}} & \multicolumn{1}{c|}{\textbf{1}} & \multicolumn{1}{c|}{\textbf{0.8}} & \textbf{0.6} \\ \hline
\multicolumn{1}{|c|}{\multirow{15}{*}{\textbf{Occurrence}}} & \multirow{5}{*}{\textbf{ST}}       & \multicolumn{1}{l|}{\textbf{Test}}       & \multicolumn{1}{c|}{0.555}      & \multicolumn{1}{c|}{0.511}        & \multicolumn{1}{c|}{0.497}        & \multicolumn{1}{c|}{0.714}      & \multicolumn{1}{c|}{0.724}        & \multicolumn{1}{c|}{0.702}        & \multicolumn{1}{c|}{-}          & \multicolumn{1}{c|}{-}            & -            \\
\hline
\end{tabular}
\caption{Overall performances of Llama3-8b (Macro F1)}
\label{tab:llama3-8b-f1}
\end{table*}

\begin{table*}[]\small
\centering
\begin{tabular}{|cc|lccccccccc|}
\hline
\multicolumn{1}{|l|}{\textbf{}}                             & \multicolumn{1}{l|}{\textbf{}}     & \multicolumn{10}{c|}{\textbf{A2R - Accuracy}}                                                                                                                                                                                                                                                                                                     \\ \hline
\multicolumn{2}{|c|}{\textbf{Datasets}}                                                          & \multicolumn{1}{l|}{\textbf{}}           & \multicolumn{3}{c|}{\textbf{Yelp}}                                                                      & \multicolumn{3}{c|}{\textbf{Go Emotions}}                                                               & \multicolumn{3}{c|}{\textbf{Beer}}                                                 \\ \hline
\multicolumn{2}{|c|}{\textbf{Shortcut}}                                                          & \multicolumn{1}{l|}{\textbf{$\lambda$}}     & \multicolumn{1}{c|}{\textbf{1}} & \multicolumn{1}{c|}{\textbf{0.8}} & \multicolumn{1}{c|}{\textbf{0.6}} & \multicolumn{1}{c|}{\textbf{1}} & \multicolumn{1}{c|}{\textbf{0.8}} & \multicolumn{1}{c|}{\textbf{0.6}} & \multicolumn{1}{c|}{\textbf{1}} & \multicolumn{1}{c|}{\textbf{0.8}} & \textbf{0.6} \\ \hline
\multicolumn{1}{|c|}{\multirow{15}{*}{\textbf{Occurrence}}} & \multirow{5}{*}{\textbf{ST}}       & \multicolumn{1}{l|}{\textbf{Test}}       & \multicolumn{1}{c|}{0.533}      & \multicolumn{1}{c|}{0.517}        & \multicolumn{1}{c|}{0.507}        & \multicolumn{1}{c|}{0.690}      & \multicolumn{1}{c|}{0.660}        & \multicolumn{1}{c|}{0.657}        & \multicolumn{1}{c|}{-}          & \multicolumn{1}{c|}{-}            & -            \\
\hline
\end{tabular}
\caption{Overall performances of A2R (Accuracy)}
\label{tab:a2r-acc}
\end{table*}

\begin{table*}[]\small
\centering
\begin{tabular}{|cc|lccccccccc|}
\hline
\multicolumn{1}{|l|}{\textbf{}}                             & \multicolumn{1}{l|}{\textbf{}}     & \multicolumn{10}{c|}{\textbf{A2R - Macro F1}}                                                                                                                                                                                                                                                                                                     \\ \hline
\multicolumn{2}{|c|}{\textbf{Datasets}}                                                          & \multicolumn{1}{l|}{\textbf{}}           & \multicolumn{3}{c|}{\textbf{Yelp}}                                                                      & \multicolumn{3}{c|}{\textbf{Go Emotions}}                                                               & \multicolumn{3}{c|}{\textbf{Beer}}                                                 \\ \hline
\multicolumn{2}{|c|}{\textbf{Shortcut}}                                                          & \multicolumn{1}{l|}{\textbf{$\lambda$}}     & \multicolumn{1}{c|}{\textbf{1}} & \multicolumn{1}{c|}{\textbf{0.8}} & \multicolumn{1}{c|}{\textbf{0.6}} & \multicolumn{1}{c|}{\textbf{1}} & \multicolumn{1}{c|}{\textbf{0.8}} & \multicolumn{1}{c|}{\textbf{0.6}} & \multicolumn{1}{c|}{\textbf{1}} & \multicolumn{1}{c|}{\textbf{0.8}} & \textbf{0.6} \\ \hline
\multicolumn{1}{|c|}{\multirow{15}{*}{\textbf{Occurrence}}} & \multirow{5}{*}{\textbf{ST}}       & \multicolumn{1}{l|}{\textbf{Test}}       & \multicolumn{1}{c|}{0.536}      & \multicolumn{1}{c|}{0.517}        & \multicolumn{1}{c|}{0.505}        & \multicolumn{1}{c|}{0.530}      & \multicolumn{1}{c|}{0.462}        & \multicolumn{1}{c|}{0.492}        & \multicolumn{1}{c|}{-}          & \multicolumn{1}{c|}{-}            & -            \\
\hline
\end{tabular}
\caption{Overall performances of A2R (Macro F1)}
\label{tab:a2r-f1}
\end{table*}

\begin{table*}[]\small
\centering
\begin{tabular}{|cc|lccccccccc|}
\hline
\multicolumn{1}{|l|}{\textbf{}}                             & \multicolumn{1}{l|}{\textbf{}}     & \multicolumn{10}{c|}{\textbf{CR - Accuracy}}                                                                                                                                                                                                                                                                                                      \\ \hline
\multicolumn{2}{|c|}{\textbf{Datasets}}                                                          & \multicolumn{1}{l|}{\textbf{}}           & \multicolumn{3}{c|}{\textbf{Yelp}}                                                                      & \multicolumn{3}{c|}{\textbf{Go Emotions}}                                                               & \multicolumn{3}{c|}{\textbf{Beer}}                                                 \\ \hline
\multicolumn{2}{|c|}{\textbf{Shortcut}}                                                          & \multicolumn{1}{l|}{\textbf{$\lambda$}}     & \multicolumn{1}{c|}{\textbf{1}} & \multicolumn{1}{c|}{\textbf{0.8}} & \multicolumn{1}{c|}{\textbf{0.6}} & \multicolumn{1}{c|}{\textbf{1}} & \multicolumn{1}{c|}{\textbf{0.8}} & \multicolumn{1}{c|}{\textbf{0.6}} & \multicolumn{1}{c|}{\textbf{1}} & \multicolumn{1}{c|}{\textbf{0.8}} & \textbf{0.6} \\ \hline
\multicolumn{1}{|c|}{\multirow{15}{*}{\textbf{Occurrence}}} & \multirow{5}{*}{\textbf{ST}}       & \multicolumn{1}{l|}{\textbf{Test}}       & \multicolumn{1}{c|}{0.555}      & \multicolumn{1}{c|}{0.540}        & \multicolumn{1}{c|}{0.540}        & \multicolumn{1}{c|}{0.693}      & \multicolumn{1}{c|}{0.752}        & \multicolumn{1}{c|}{0.762}        & \multicolumn{1}{c|}{-}          & \multicolumn{1}{c|}{-}            & -            \\
\hline
\end{tabular}
\caption{Overall performances of CR (Accuracy)}
\label{tab:cr-acc}
\end{table*}

\begin{table*}[]\small
\centering
\begin{tabular}{|cc|lccccccccc|}
\hline
\multicolumn{1}{|l|}{\textbf{}}                             & \multicolumn{1}{l|}{\textbf{}}     & \multicolumn{10}{c|}{\textbf{CR-Macro F1}}                                                                                                                                                                                                                                                                                                        \\ \hline
\multicolumn{2}{|c|}{\textbf{Datasets}}                                                          & \multicolumn{1}{l|}{\textbf{}}           & \multicolumn{3}{c|}{\textbf{Yelp}}                                                                      & \multicolumn{3}{c|}{\textbf{Go Emotions}}                                                               & \multicolumn{3}{c|}{\textbf{Beer}}                                                 \\ \hline
\multicolumn{2}{|c|}{\textbf{Shortcut}}                                                          & \multicolumn{1}{l|}{\textbf{$\lambda$}}     & \multicolumn{1}{c|}{\textbf{1}} & \multicolumn{1}{c|}{\textbf{0.8}} & \multicolumn{1}{c|}{\textbf{0.6}} & \multicolumn{1}{c|}{\textbf{1}} & \multicolumn{1}{c|}{\textbf{0.8}} & \multicolumn{1}{c|}{\textbf{0.6}} & \multicolumn{1}{c|}{\textbf{1}} & \multicolumn{1}{c|}{\textbf{0.8}} & \textbf{0.6} \\ \hline
\multicolumn{1}{|c|}{\multirow{15}{*}{\textbf{Occurrence}}} & \multirow{5}{*}{\textbf{ST}}       & \multicolumn{1}{l|}{\textbf{Test}}       & \multicolumn{1}{c|}{0.533}      & \multicolumn{1}{c|}{0.510}        & \multicolumn{1}{c|}{0.516}        & \multicolumn{1}{c|}{0.483}      & \multicolumn{1}{c|}{0.589}        & \multicolumn{1}{c|}{0.508}        & \multicolumn{1}{c|}{-}          & \multicolumn{1}{c|}{-}            & -            \\
\hline
\end{tabular}
\caption{Overall performances of CR (Macro F1)}
\label{tab:cr-f1}
\end{table*}

\begin{table*}[]\small
\centering
\begin{tabular}{|cc|lccccccccc|}
\hline
\multicolumn{1}{|l|}{\textbf{}}                             & \multicolumn{1}{l|}{\textbf{}}     & \multicolumn{10}{c|}{\textbf{AFR - Accuracy}}                                                                                                                                                                                                                                                                                                     \\ \hline
\multicolumn{2}{|c|}{\textbf{Datasets}}                                                          & \multicolumn{1}{l|}{\textbf{}}           & \multicolumn{3}{c|}{\textbf{Yelp}}                                                                      & \multicolumn{3}{c|}{\textbf{Go Emotions}}                                                               & \multicolumn{3}{c|}{\textbf{Beer}}                                                 \\ \hline
\multicolumn{2}{|c|}{\textbf{Shortcut}}                                                          & \multicolumn{1}{l|}{\textbf{$\lambda$}}     & \multicolumn{1}{c|}{\textbf{1}} & \multicolumn{1}{c|}{\textbf{0.8}} & \multicolumn{1}{c|}{\textbf{0.6}} & \multicolumn{1}{c|}{\textbf{1}} & \multicolumn{1}{c|}{\textbf{0.8}} & \multicolumn{1}{c|}{\textbf{0.6}} & \multicolumn{1}{c|}{\textbf{1}} & \multicolumn{1}{c|}{\textbf{0.8}} & \textbf{0.6} \\ \hline
\multicolumn{1}{|c|}{\multirow{15}{*}{\textbf{Occurrence}}} & \multirow{5}{*}{\textbf{ST}}       & \multicolumn{1}{l|}{\textbf{Test}}       & \multicolumn{1}{c|}{0.641}      & \multicolumn{1}{c|}{0.634}        & \multicolumn{1}{c|}{0.628}        & \multicolumn{1}{c|}{0.902}      & \multicolumn{1}{c|}{0.900}        & \multicolumn{1}{c|}{0.873}        & \multicolumn{1}{c|}{-}          & \multicolumn{1}{c|}{-}            & -            \\
\hline
\end{tabular}
\caption{Overall performances of AFR (Accuracy)}
\label{tab:afr-acc}
\end{table*}

\begin{table*}[]\small
\centering
\begin{tabular}{|cc|lccccccccc|}
\hline
\multicolumn{1}{|l|}{\textbf{}}                             & \multicolumn{1}{l|}{\textbf{}}     & \multicolumn{10}{c|}{\textbf{AFR - Macro F1}}                                                                                                                                                                                                                                                                                                     \\ \hline
\multicolumn{2}{|c|}{\textbf{Datasets}}                                                          & \multicolumn{1}{l|}{\textbf{}}           & \multicolumn{3}{c|}{\textbf{Yelp}}                                                                      & \multicolumn{3}{c|}{\textbf{Go Emotions}}                                                               & \multicolumn{3}{c|}{\textbf{Beer}}                                                 \\ \hline
\multicolumn{2}{|c|}{\textbf{Shortcut}}                                                          & \multicolumn{1}{l|}{\textbf{$\lambda$}}     & \multicolumn{1}{c|}{\textbf{1}} & \multicolumn{1}{c|}{\textbf{0.8}} & \multicolumn{1}{c|}{\textbf{0.6}} & \multicolumn{1}{c|}{\textbf{1}} & \multicolumn{1}{c|}{\textbf{0.8}} & \multicolumn{1}{c|}{\textbf{0.6}} & \multicolumn{1}{c|}{\textbf{1}} & \multicolumn{1}{c|}{\textbf{0.8}} & \textbf{0.6} \\ \hline
\multicolumn{1}{|c|}{\multirow{15}{*}{\textbf{Occurrence}}} & \multirow{5}{*}{\textbf{ST}}       & \multicolumn{1}{l|}{\textbf{Test}}       & \multicolumn{1}{c|}{0.638}      & \multicolumn{1}{c|}{0.631}        & \multicolumn{1}{c|}{0.625}        & \multicolumn{1}{c|}{0.826}      & \multicolumn{1}{c|}{0.812}        & \multicolumn{1}{c|}{0.781}        & \multicolumn{1}{c|}{-}          & \multicolumn{1}{c|}{-}            & -            \\
\hline
\end{tabular}
\caption{Overall performances of AFR (Macro F1)}
\label{tab:afr-f1}
\end{table*}

\end{document}